\documentclass[journal]{IEEEtran}
\usepackage{amsmath,amsfonts}
\usepackage{algorithmic}
\usepackage{algorithm}
\usepackage[table]{xcolor}
\usepackage{colortbl}
\usepackage{booktabs}
\usepackage{geometry}
\usepackage{array}
\usepackage[caption=false,font=normalsize,labelfont=sf,textfont=sf]{subfig}
\usepackage{textcomp}
\usepackage{stfloats}
\usepackage{url}
\usepackage{verbatim}
\usepackage{graphicx}
\usepackage{cite}
\usepackage{colortbl} 
\usepackage{xcolor}

\definecolor{revisedcolor}{RGB}{0,0,0} 
\newcommand{\myrevisedcolor}[1]{\textcolor{revisedcolor}{#1}}

\begin{document}

\title{EvaNet: Towards More Efficient and Consistent Infrared and Visible Image Fusion Assessment
}

\author{Chunyang Cheng,
        Tianyang Xu,
        Xiao-Jun Wu,
        Tao Zhou,
        Hui Li,
        Zhangyong Tang, and 
        Josef Kittler

\IEEEcompsocitemizethanks{\IEEEcompsocthanksitem 
C. Cheng and Z. Tang are with the Wuxi School of Medicine and also with the School of Food Science and Technology, Jiangnan University, Wuxi 214122, P.R. China. 
T. Xu, X.-J. Wu, T. Zhou and H. Li are with the School of Artificial Intelligence and Computer Science, Jiangnan University, Wuxi 214122, P.R. China. 
(Corresponding author: X.-J. Wu, e-mail: wu\_xiaojun@jiangnan.edu.cn)
\IEEEcompsocthanksitem Josef Kittler is with the Centre for Vision, Speech and Signal Processing, University of Surrey, Guildford GU2 7XH, U.K. (e-mail: j.kittler@surrey.ac.uk)
}
\thanks{Manuscript received July, 2025.}}

\markboth{Preprint for IEEE TPAMI, July~2025}%
{Shell \MakeLowercase{\textit{et al.}}: A Sample Article Using IEEEtran.cls for IEEE Journals}


\maketitle

\begin{abstract}
Evaluation is essential in image fusion research, yet most existing metrics are directly borrowed from other vision tasks without proper adaptation. These traditional metrics, often based on complex image transformations, not only fail to capture the true quality of the fusion results but also are computationally demanding.
To address these issues, we propose a unified evaluation framework specifically tailored for image fusion. At its core is a lightweight network designed efficiently to approximate widely used metrics, following a divide-and-conquer strategy.
Unlike conventional approaches that directly assess similarity between fused and source images, we first decompose the fusion result into infrared and visible components.
The evaluation model is then used to measure the degree of information preservation in these separated components, effectively disentangling the fusion evaluation process.
During training, we incorporate a contrastive learning strategy and inform our evaluation model by  perceptual scene assessment provided by a large language model.
Last, we propose the first consistency evaluation framework, which measures the alignment between image fusion metrics and human visual perception, using both independent no-reference scores and downstream tasks performance as objective references.
Extensive experiments show that our learning-based evaluation paradigm delivers both superior efficiency (up to 1,000 times faster) and greater consistency across a range of standard image fusion benchmarks.
\myrevisedcolor{Our code will be publicly available at \url{https://github.com/AWCXV/EvaNet}.}

\end{abstract}


\section{Introduction}
\IEEEPARstart{I}mage fusion is a technique to combine the information from different modalities or images with different settings to produce a single fused image that is more consistent with the human visual perception and helps to improve the downstream vision tasks performance~\cite{zhang2023IVIFsurveyPAMI}.
This technique has been widely used in different areas, such as remote sensing, video monitoring, and medical diagnosis~\cite{karim2023fusionSurveyInfFus}.

Among the many fusion scenarios, Infrared and Visible Image Fusion (IVIF) has attracted considerable attention due to the complementarity of the imaging modalities~\cite{lihuafeng2025mulfs,deng2024mmdrfuse,ma2025s4fusion}.
Infrared images can effectively highlight thermal targets, making them robust to poor lighting and adverse weather conditions, while visible images preserve rich texture and structural details of the scene.
Fusing these two modalities into a single image allows for enhanced scene understanding and significantly benefits high-level tasks such as object detection and semantic segmentation.

\begin{figure}[t]
\centering
\includegraphics[width=0.95\linewidth]{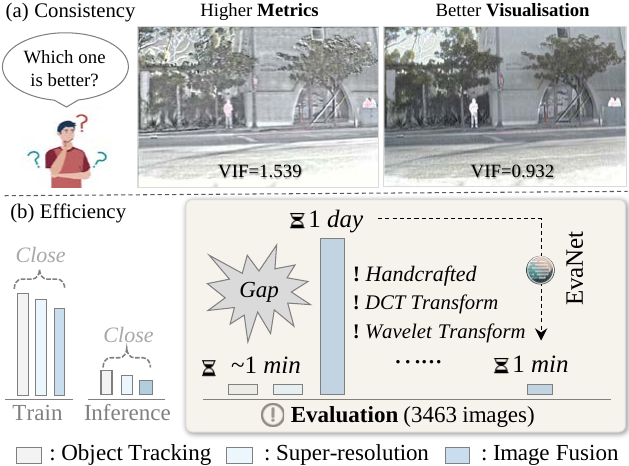}
\caption{
\myrevisedcolor{
An illustration of the consistency and the efficiency issues (sub-figure (a) and (b)) of existing image fusion metrics.
Image fusion evaluation relies heavily on traditional signal processing techniques (\textit{e.g.}, discrete cosine transform (DCT) and wavelet transform) or metrics adapted from other vision tasks.
Without appropriate adjustment, these metrics lack consistency, \textit{i.e.}, better visualisation cannot always correspond higher metric values.
Moreover, the evaluation phase in current image fusion settings incurs high processing time costs.
The proposed EvaNet addresses these issues effectively by significantly reducing the evaluation time and enhancing consistency.}
}
\label{figure_imbalance_training_inf_eva}
\end{figure}

Recently, we have witnessed rapid advances in image fusion methods driven by deep learning.
Sophisticated frameworks such as GCN~\cite{li2023gcnfusion}, GAN~\cite{liu2022target}, Transformer~\cite{cheng2025gifnet}, Mamba~\cite{xie2024fusionmamba}, and Diffusion Models~\cite{zhu2025cvprMultispecDiffusion} have been introduced to address various challenges in the fusion process.
However, most existing research has focused on the training and inference stages, \textit{i.e.}, generating the fused image, while the evaluation of fusion quality remains relatively unexplored, especially in IVIF tasks.

Although much effort has been devoted to improving fusion outcomes~\cite{cheng2025fusionbooster,zhang2025omnifuse,25TPAMI_freefusion}, a fundamental question remains: \textit{Are these improvements guided by the most appropriate optimisation direction?}
The answer critically depends on the reliability and consistency of the evaluation metrics used.
Unfortunately, current fusion assessments largely adopt metrics from other vision tasks, such as super-resolution or denoising, which typically rely on ground-truth (GT) references that IVIF lacks.
To circumvent this, many works simply treat the fusion result as a pseudo-GT and compute similarity between the fused image and each source image. This practice, however, violates the original design of these metrics, which are intended for comparing images of the same modality.
Additionally, fusion tasks inherently involve multiple inputs from different modalities, which should not necessarily  contribute equally to the fusion process, but instead, their weight should reflect the scene context. The current metrics fail to consider this natural requirement and they suffer from the issue of insufficient consistency (Fig.~\ref{figure_imbalance_training_inf_eva} (a)).

\begin{figure*}[t]
\centering
\includegraphics[width=0.7\linewidth]{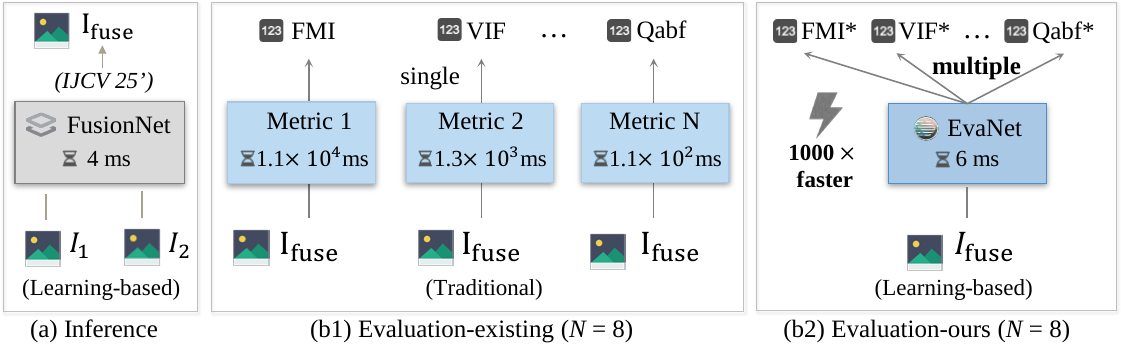}
\caption{
\myrevisedcolor{
The speed (milliseconds per image) imbalance between inference and evaluation in image fusion.
Traditional metrics rely on separate, computationally intensive procedures to perform different assessments, significantly slowing down the evaluation phase.
In contrast, the proposed EvaNet generates multiple evaluation results simultaneously within a single forward pass, offering acceleration by a factor of up to 1000, and as will be demonstrated, also considerably improved metric consistency.}
}
\label{figure_imbalance_inf_eva_detail}
\end{figure*}

Another critical limitation lies in evaluation efficiency.
Generally, it takes much more time to train a specific vision model than evaluating the task performance~\cite{tang2024generative,deng2024coconut,xu2024learning}.
However, as illustrated in Fig.~\ref{figure_imbalance_training_inf_eva} (b), although image fusion and super-resolution have comparable time consumption during training and inference phases, the evaluation phase is disproportionately slower.
For example, assessing the full LLVIP test set~\cite{jia2021llvip} (3,463 images) using traditional fusion metrics can take nearly 24 hours, while the evaluation for super-resolution tasks may finish in under a minute. 
This leads to a significant imbalance in the overall pipeline, where the evaluation phase becomes a bottleneck.
As a consequence, researchers often evaluate their methods on only a subset of the test dataset (\textit{e.g.}, 10\%), limiting the statistical reliability of the reported results.

To tackle these challenges, we propose a unified learning-based evaluation framework.
First, we design a lightweight network, EvaNet,  efficiently to approximate widely-used fusion metrics. 
Built on a divide-and-conquer strategy, EvaNet disentangles the evaluation process by first decomposing the fused image into its infrared and visible components, allowing the network to assess the extent to which information from each modality is preserved.
Besides, the environmental conditions significantly influence the mixing bias between source images during fusion, which is often overlooked in existing evaluation methods.
Thus, the idea of contrastive learning, informed by a perceptual assessment of the scene environment furnished by a large language model (LLM), is introduced as the key innovative concept during training to help enhancing the consistency of the delivered metrics.

To improve the runtime efficiency and to handle diverse metric types, EvaNet adopts a multi-head output architecture that simultaneously predicts multiple metrics.
As shown in Fig.~\ref{figure_imbalance_inf_eva_detail}, although fusion inference is typically fast, traditional metrics require separate executions for each assessment, which is computationally expensive and time-consuming.
In contrast, our proposed EvaNet produces multiple evaluation results simultaneously in a single forward pass, dramatically narrowing the gap between inference and evaluation time.


Finally, while prior efforts such as~\cite{liu2024tpamiMetricMFIF} have examined the consistency of metrics in multi-focus image fusion (where GT is available), no such consistency study exists for IVIF.
We address this gap by introducing a Metric Consistency (MC) measurement based on the average ranking error between the image quality assessment and the detection or segmentation model grounded by third-party references.

The main contributions of this work are as follows:
\begin{itemize}
\item We propose EvaNet, the first learning-based framework to compute traditional image fusion metrics in the IVIF context with improved evaluation consistency.
\item We develop a lightweight multi-head architecture which, thanks to contrastive learning, enables one-pass estimation of multiple quality metrics and achieves over 1,000× acceleration.
\item A fine-grained evaluation mechanism is developed by analysing the modality-specific components, taking into account the perceptual assessment of the scene provided by LLM, enabling a more accurate weighting of the modality contributions.
\item We introduce a third-party-supported consistency evaluation protocol tailored for IVIF, providing a principled foundation for future research on metric reliability.
\item Extensive experiments on multiple benchmarks validate the superiority of EvaNet in terms of accuracy, consistency, and efficiency.
\end{itemize}




\section{Related Works}
An overview of recent state-of-the-art image fusion algorithms and evaluation metrics are presented in Table~\ref{statistics_sota_algorithm_metrics}.

\subsection{Image Fusion Algorithms}
Over the years, image fusion algorithms have adopted a diverse range of technical frameworks.
While the boundaries may sometimes overlap, the techniques can broadly be categorised according to their core approach into five main groups: traditional signal processing methods, CNN-based approaches, Transformer-based architectures, Mamba-based models, and Diffusion-based generative methods.

\begin{table}[tbp]
  \centering

  \caption{Statistics of recent Infrared and Visible Image Fusion (IVIF) and Multi-Focus Image Fusion (MFIF) algorithms and metrics.}
  
  \resizebox{1\linewidth}{!}{
    \begin{tabular}{cccc}
    \hline
    Category & IVIF Algorithm & IVIF Metric & MFIF Metric \\
    \hline
    Traditional & ~\cite{li2023lrrnet},~\cite{li2020mdlatlrr},~\cite{wang2014multiscal2}… & \cite{cheng2023textfusion}, Qabf, VIF… & SSIM, PSNR, … \\
    CNN   & \cite{cheng2025gifnet}, \cite{zhang2024mrfs},\cite{li2018densefuse}… & \textbf{Unexplored} & 2024' TPAMI~\textbf{\cite{liu2024tpamiMetricMFIF}} \\
    Transformer &  \cite{Bai_2025_task_learnable},\cite{Wu_2025_CVPR_sam_DROP}, \cite{liu2025cvpr_dcevo}… & -     & - \\
    Mamba & \cite{xie2024fusionmamba}, \cite{li2024mambadfuse}… & -     & - \\
    Diffusion &  \cite{Zhao_2023_ICCV_DDFM},\cite{yi2024diff}… & -     & - \\
    \hline
    \end{tabular}%
    }
  \label{statistics_sota_algorithm_metrics}%
\end{table}%

\textit{Traditional Methods}.
Conventional image fusion methods rely primarily on classical signal processing techniques such as multi-scale decomposition~\cite{wang2014multiscal2} and low-rank representation~\cite{li2020mdlatlrr} to extract handcrafted features or representations from the source images.
Manually designed fusion strategies are then utilised to aggregate these features or representations, followed by reconstruction procedures to generate the final fused image.

\textit{CNN-based Methods.}
To overcome the limitations of handcrafted features~\cite{li2018densefuse} and heuristic fusion strategies~\cite{xu2020u2fusion}, a variety of CNN-based approaches have been proposed.
MUFusion~\cite{cheng2023mufusion} introduces a memory unit to enhance the learning capacity of CNNs.
Additionally, several methods integrate downstream tasks into training to inject semantic awareness into the fusion process~\cite{liu2022target, Zhao2023metafusion,zhang2024mrfs,zhang2025omnifuse}, thereby improving performance in detection and segmentation.

\textit{Transformer-based Methods.}
Transformers have gained popularity in image fusion due to their ability to model long-range dependencies and global context~\cite{rao2022tgfuse}.
For instance, SwinFusion~\cite{ma2022swinfusion} demonstrates the effectiveness of hierarchical Transformer structures in fusion tasks.
CDDFuse~\cite{zhao2023cddfuse} adopts the Restormer~\cite{zamir2022restormer} backbone to further improve imaging quality of the fusion results.
Transformers also show promise in joint fusion-segmentation frameworks, such as SegMiF~\cite{liu2023fusion_seg_ICCV2023}.

\textit{Mamba-based Methods.}
Mamba is a structured state space model (SSM)-based module known for its computational efficiency in capturing global dependencies.
Recent work has adapted Mamba for image fusion~\cite{xie2024fusionmamba, li2024mambadfuse}, highlighting its advantages over traditional Transformer models in certain scenarios.
LE-Mamba~\cite{cao2024lemamba} further incorporates spectral information through state sharing, achieving competitive performance on multispectral and panchromatic image fusion tasks.

\textit{Diffusion-based Methods.}
Denoising Diffusion Probabilistic Models (DDPMs)\cite{ho2020denoising} offer a powerful generative framework capable of synthesising high-fidelity images via iterative denoising.
In DDFM~\cite{Zhao_2023_ICCV_DDFM}, the authors use the generative strength and stability of DDPMs to model the multi-modal image fusion process as a conditional sampling problem.
Through the integration of diffusion sampling and a statistical inference, it achieves high-quality fusion results across infrared and visible image fusion benchmarks.
Extensions of this paradigm incorporate prior constraints and task-oriented guidance~\cite{yi2024diff, zhu2025cvprMultispecDiffusion}, further improving robustness and fidelity.

From the above discussion, it is clear that image fusion algorithms have evolved significantly through the integration of diverse technical paradigms.
However, despite the notable performance gains achieved by these methods, the importance of selecting reliable and consistent evaluation metrics remains largely overlooked within the community.

\myrevisedcolor{\subsection{Traditional Image Fusion Evaluation}}
In the context of image fusion evaluation, existing assessment strategies share similar limitations with traditional image fusion algorithms. These metrics typically rely on classical signal processing techniques to compute numerical scores based on handcrafted features.

In this work, we primarily focus on improving weighted-average evaluation metrics.
Such metrics are usually borrowed from other vision tasks or derived from gradient and edge information, with fixed weighting ratios that lack awareness of the actual fusion scenario.
Fundamentally, they can be formulated as:
\begin{equation}
    RQ_f = w_{\textrm{ir}}*Q(I_{\textrm{ir}},I_f) + w_{\textrm{vis}}*Q(I_{\textrm{vis}},I_f),
\end{equation}
where $Q$ denotes a reference-based quality metric, such as SSIM, FMI, or VIF. 
$w_{\textrm{ir}}$ and $w_{\textrm{vis}}$ represent weighting factors associated with the infrared and visible modalities, respectively.
In most cases, these are set equally, \textit{e.g.}, $w_{\textrm{ir}}=w_{\textrm{vis}}=1$, implying that both modalities are assumed to contribute equally to the final quality score.
This approach enables researchers to evaluate the results using metrics that reflect structural similarity (SSIM~\cite{ma2015metricmssim}), mutual information (FMI~\cite{haghighat2011FMI}), visual information fidelity (VIF~\cite{sheikh2006VIF}), correlation (CC), and pixel-level similarity (PSNR). 
However, such a direct adaptation fails to consider the environmental context of the fusion task, and is thus unable to accurately reflect the relative contribution of each modality in different scenarios.
To overcome these limitations, our proposed EvaNet replaces handcrafted features with adaptive deep features tailored for different fusion metrics.
In addition, a large language model (LLM) is used to construct an environment-aware perception module, which enables the evaluation framework to weigh the modality contributions more appropriately.

\myrevisedcolor{\subsection{Learning-based Image Fusion Evaluation}}
Recent research has aimed to improve the resulting image assessment in the multi-focus image fusion (MFIF) domain~\cite{liu2024tpamiMetricMFIF}.
However, this work is tailored to tasks where Ground-Truth (GT) images are available, and thus cannot directly be applied to infrared and visible image fusion (IVIF), where GT is typically absent.
In TextFusion~\cite{cheng2023textfusion}, the authors attempt to overcome the limitations of fixed-weight traditional metrics by introducing enhanced versions of CC, SSIM, and Qabf, collectively referred to as textual attention metrics, to improve evaluation consistency.
Nonetheless, these improvements overlook computational efficiency: the metrics remain dependent on multiple separate executions of the original assessments, which requires substantial time and resource costs.
Besides, as highlighted in Table~\ref{statistics_sota_algorithm_metrics}, advanced learning-based techniques such as convolutional neural networks (CNNs) have not yet been fully explored within the IVIF evaluation domain.
In this work, we propose a learning-based framework that systematically improves the consistency of existing image fusion assessments.
Our approach emphasises both comprehensive evaluation coverage and practical efficiency by delivering multiple metric predictions through a single forward pass network, thereby streamlining the evaluation process for IVIF research.

\subsection{Component Analysis}
FusionBooster~\cite{cheng2025fusionbooster} introduces a general post-processing framework designed to enhance arbitrary image fusion methods.
Within this framework, an autoencoder-based model, \textit{i.e.}, information probe, is used to decompose the fused image into its constituent components, \textit{e.g.}, infrared and visible modalities.
Inspired by this design, our proposed EvaNet incorporates a similar probe mechanism to facilitate fine-grained assessment of fusion results.
By analysing the contribution of each modality separately, EvaNet enables more consistent and interpretable evaluations across a variety of fusion metrics.



\begin{figure}[t]
\centering
\includegraphics[width=1\linewidth]{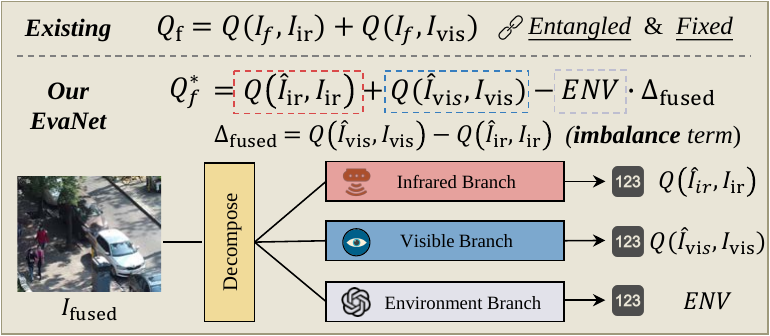}
\caption{
\myrevisedcolor{
Overview of the proposed EvaNet framework.
Our method replaces traditional image fusion assessment processes by a lightweight learning-based network to significantly improve evaluation efficiency.
In addition, a divide-and-conquer strategy is used to disentangle and independently measure the information preserved from each source modality.
The environment branch, as part of the three-branch design, introduces an adaptive penalty mechanism to mitigate modality imbalance in challenging fusion scenarios.}
}
\label{figure_framework}
\end{figure}

\section{The proposed EvaNet}
\subsection{Framework}

Conventional image fusion evaluation primarily relies on image quality assessments that directly measure the overall information transferred from single-modality inputs to the fused output.
This entangled evaluation approach (top section of Fig.~\ref{figure_framework}) often conceals potential fusion issues, as it does not disentangle the contributions of each modality.
Furthermore, the weighting of infrared and visible modalities is typically fixed and treated as equally important, regardless of varying scene conditions.

To address these limitations, we propose a decomposition-based evaluation framework (bottom section of Fig.~\ref{figure_framework}) that enables a more interpretable and fine-grained assessment.
The framework consists of three main branches: an infrared component branch, a visible component branch, and an environment-awareness branch.
The idea of decomposition may sound counter-intuitive. However, by incorporating a decomposition operation, we are able to separately evaluate specific quality metrics (\textit{e.g.}, structural similarity) between the disentangled fusion components ($\hat{I}_{\textrm{ir}}$, $\hat{I}_{\textrm{vis}}$) and their corresponding source images ($I_{\textrm{ir}}$, $I_{\textrm{vis}}$).
This design allows for a clearer understanding of how much information each modality contributes to the final fusion result.

To mitigate the lack of environmental context in traditional assessments, the EvaNet incorporates an environment branch to introduce an adaptive weighting factor $ENV$, which adjusts the final evaluation score to eliminate modality imbalance.
For instance, in poor visibility conditions, the metric score of the visible light modality will be reduced accordingly. 
The final assessment is formulated as:
\begin{equation}
\label{eqQstar}    Q^*=Q(\hat{I}_{\textrm{ir}},I_{\textrm{ir}})+Q(\hat{I}_{\textrm{vis}},I_{\textrm{vis}})-ENV\cdot \Delta,
\end{equation}
\myrevisedcolor{where $\Delta$ denotes the modality imbalance term, defined as the difference between the visible and infrared quality scores, i.e., $Q(\hat{I}_{\textrm{vis}}, I_{\textrm{vis}}) - Q(\hat{I}_{\textrm{ir}}, I_{\textrm{ir}})$.}
A higher $\Delta$ indicates that the fusion result retains more information from the visible modality than the infrared one.
\myrevisedcolor{By introducing the environment-aware factor $ENV$, the penalty imposed by $\Delta$ is adaptively modulated according to scene conditions, such that visible-dominant imbalance is penalised more strongly under adverse visibility or illumination.}

\myrevisedcolor{We emphasise that Eq.~\eqref{eqQstar} does not redefine the original metric $Q$, but provides a consistency-oriented adapted scoring built upon fast surrogate predictions of existing metric definitions.
}

\begin{figure*}[t]
\centering
\includegraphics[width=1\linewidth]{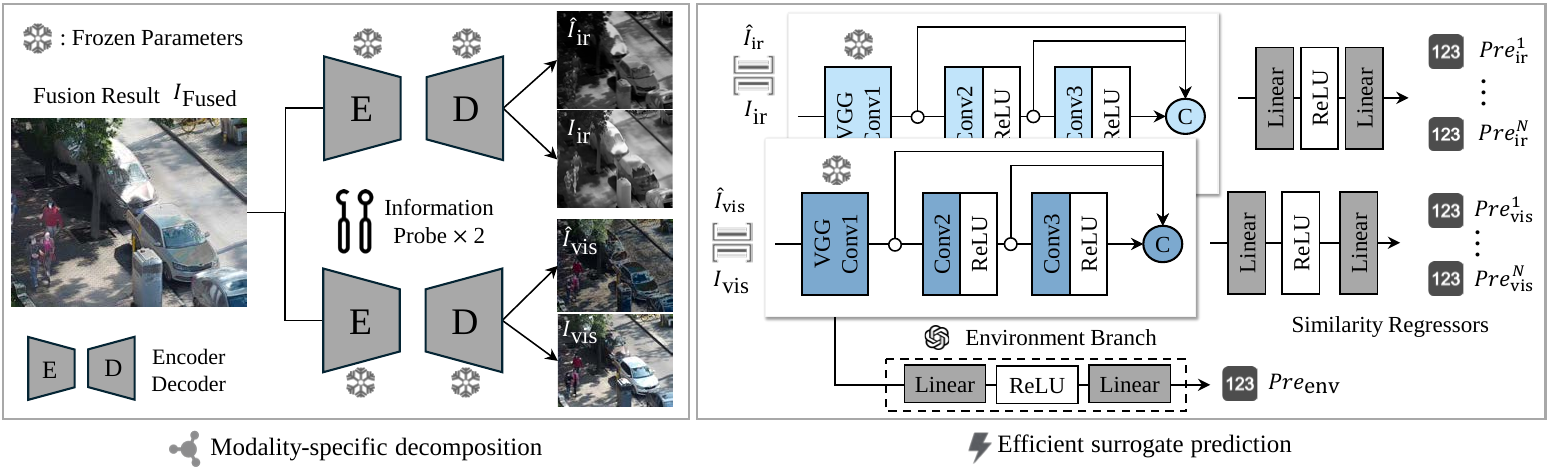}
\caption{
\myrevisedcolor{The network architecture of the proposed EvaNet (see Sec.~\ref{sec_network}).
The model consists of two main components.
The left part shows the modality-specific decomposition, implemented using two lightweight information probes~\cite{cheng2025fusionbooster}.
The right part illustrates the surrogate metric prediction process, consisting of three branches: two modality branches correspond to the infrared and visible inputs, while the environment branch, guided by a large language model (LLM), predicts an adaptive weighting factor ($ENV$) to eliminate any modality imbalance during evaluation.}
}
\label{figure_networkArchitecture}
\end{figure*}

\subsection{Network Architecture}
\label{sec_network}
The network architecture of EvaNet is shown in Fig.\ref{figure_networkArchitecture}.
Inspired by FusionBooster\cite{cheng2025fusionbooster}, we adopt an information probe to estimate the information preserved from each source modality in the fused image.
In our design, two lightweight pre-trained information probes (each requiring only 40KB of parameters) are used to decompose the fused image into its infrared and visible components, enabling a fine-grained evaluation without significant computational overhead.
This process can be formulated as:

\begin{equation}
    \hat{I}_{\textrm{vis}},\hat{I}_{\textrm{ir}} = IP(I_{\textrm{f}}),
\end{equation}
where $\hat{I}_{\textrm{vis}}$ and $\hat{I}_{\textrm{ir}}$ denote the visible and infrared components derived from the fusion result $I_{\textrm{f}}$ via the information probe $IP$.

For the two modality branches, we first extract perceptual features from both the original source images ($I_{\textrm{ir}}$, $I_{\textrm{vis}}$) and the corresponding decomposed components ($\hat{I}_{\textrm{ir}}$, $\hat{I}_{\textrm{vis}}$) using a pre-trained vision backbone~\cite{vgg}.
These features are then passed through two convolutional layers tailored for each modality.
Finally, two similarity regressors output $2N$ image quality scores: $Pre^{1..N}_{\textrm{ir}}$ and $Pre^{1..N}_{\textrm{vis}}$, corresponding to $N$ different evaluation metrics (\textit{e.g.}, structural similarity of the decomposed components and the original images).


\myrevisedcolor{The environment branch operates on the visible modality and is designed to provide an adaptive penalty weight $ENV$ for the modality imbalance term $\Delta$ in Eq.~\eqref{eqQstar}.}
It shares the same feature backbone and feeds into a dedicated regressor, which estimates the reliability of visible information under the current environmental conditions.

\begin{figure}[t]
\centering
\includegraphics[width=0.8\linewidth]{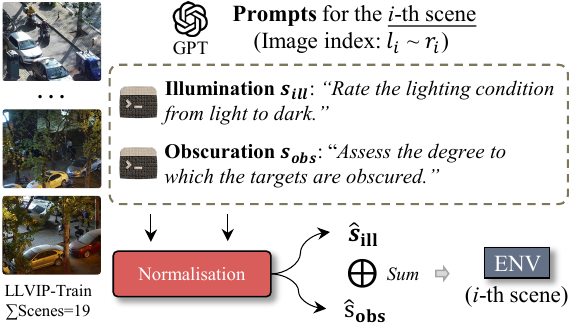}
\caption{
An illustration of perceptual scene environment assessment. 
Left: some examples showing that the LLVIP-train dataset contains multiple image pairs with limited scene diversity, where foreground objects vary but the background remains consistent.
Right: environment label generation process using a Generative Pre-trained Transformer (GPT)~\cite{OpenAI2025ChatGPT-4o}, which estimates scene conditions such as illumination and obscuration to guide the training of the environment branch in EvaNet.
}
\label{figure_env}
\end{figure}

\subsection{Perceptual Environment Assessment}
\label{sec_env}
To incorporate environmental awareness into the evaluation framework, we design an environment branch that gauges the visibility and illumination conditions of each scene, which directly influences the reliability of the visible modality content in the fusion results.
The key idea is to detect environmental degradations—such as poor lighting or visual obstructions— in visible light images. Since visible information is often vulnerable under adverse conditions, it is crucial to penalise overreliance on this modality when it is degraded.

To this end, we use the reasoning capability of large language models (LLMs), particularly ChatGPT-4o~\cite{OpenAI2025ChatGPT-4o}, to generate control signals for the environment branch.
As illustrated in Fig.~\ref{figure_env}, the training split of LLVIP contains a large number of image pairs, yet only a limited set of distinct scenes (\textit{i.e.}, scenes with the same background but varying foreground objects).
This property allows us to assign consistent environmental labels to images from the same scene.

For the $i$-th scene, we randomly select a sample from its corresponding range (from $l_i$ to $r_i$) and query ChatGPT to assess two key attributes: illumination condition ($s_{\textrm{ill}}$) and obscuration level ($s_{\textrm{obs}}$).
\myrevisedcolor{
Specifically, the LLM is prompted to rate illumination from light to dark and obscuration from lightly to heavily obscured, each on a continuous scale ranging from 0 to 0.5, following a fixed output format.
}
The exact prompt templates used to generate these attributes are shown on the right-hand side of Fig.~\ref{figure_env}.
\myrevisedcolor{
To reduce scale bias across different scenes and models, the raw scores are further normalised using min--max normalisation within the dataset, mapping both attributes into the range $[0, 0.5]$.
The resulting ENV scores are obtained by comibing both attributes together, \textit{i.e.}
\begin{equation}
ENV = \hat{s}_{\textrm{ill}} + \hat{s}_{\textrm{obs}},
\end{equation}
where $\hat{s}_{\textrm{ill}}$ and $\hat{s}_{\textrm{obs}}$ denote the normalised illumination and obscuration scores, respectively.
In this formulation, lower illumination and higher obscuration jointly indicate a more challenging visual environment, leading to a larger $ENV$ value.
}
This value is then used during training as a penalty factor to modulate the evaluation score (see Eq.~\eqref{eqQstar}), ensuring that the metric reflects not only the similarity of modalities but also their contextual reliability.

\begin{figure}[t]
\centering
\includegraphics[width=1\linewidth]{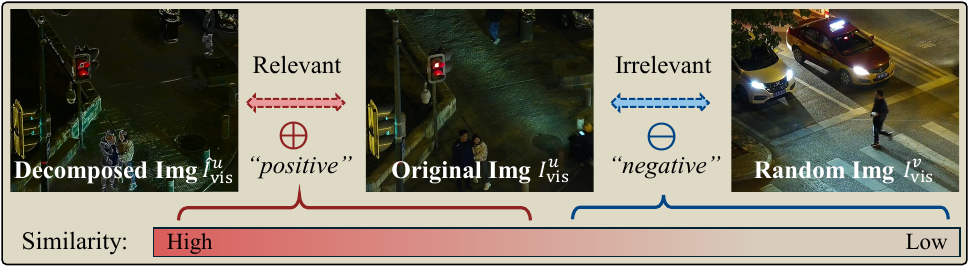}
\caption{
An overview of the training strategy based on contrastive learning.
The decomposed infrared or visible image (obtained from the fusion result, \textit{e.g.}, $\hat{I}^u_{\textrm{vis}}$) and its corresponding original input ($I^u_{\textrm{vis}}$) form a positive sample pair, guiding the model to learn the similarity range of relevant images.
In contrast, the original image paired with an unrelated randomly selected image ($I^v_{\textrm{vis}}$) forms a negative sample, encouraging the network to suppress metric predictions for unrelated content.
}
\label{figure_training}
\end{figure}

\subsection{Training}
To enable EvaNet to learn metric predictions across a wide range of value distributions, we adopt a training strategy inspired by contrastive learning.
Specifically, for each training sample, we treat the decomposed infrared or visible component (obtained from the fusion result) and its corresponding original input image as a positive pair. These pairs exhibit semantic and structural correlation, and their metric scores, computed using MATLAB implementations of standard definitions, fall within relatively high ranges. The network is supervised to predict these values accurately.

In parallel, we construct negative pairs by randomly selecting two unrelated images from the same modality.
As these image pairs are semantically disjoint, the similarity metrics between them are naturally lower.
These low scores are likewise used as supervision targets to train the network to suppress the predicted similarity values for dissimilar inputs.

This contrastive setup not only helps the model distinguish between relevant and irrelevant image content, but also avoids collapsed representations by explicitly enforcing metric separation in the learned feature space.
The overall training loss is defined as:
\begin{equation}
\mathcal{L}_{\textrm{total}} = \mathcal{L}^{\textrm{ir}}_{\textrm{met}}+\mathcal{L}^{\textrm{vis}}_{\textrm{met}}+\mathcal{L}_{\textrm{env}}.
\end{equation}
where $\mathcal{L}^{\textrm{ir}}_{\textrm{met}}$ and $\mathcal{L}^{\textrm{vis}}_{\textrm{met}}$ denote the metric regression losses for the infrared and visible branches, respectively, and $\mathcal{L}_{\textrm{env}}$ is the environment perception loss.

The metric loss for each modality branch is composed of positive and negative components:

\begin{equation}
\mathcal{L}_X^{met}=\mathcal{L}_X^{\textrm{pos}}+\mathcal{L}_X^{\textrm{neg}},
\end{equation}
\begin{equation}
\mathcal{L}^{\textrm{pos}}_X = \mathcal{L}_{\textrm{mse}}(Branch_X(I^u_{\textrm{X}},\hat{I}^u_{\textrm{X}}),M_{\textrm{pos}}),
\end{equation}
\begin{equation}
\mathcal{L}_X^{\textrm{neg}} = \mathcal{L}_{\textrm{mse}}(Branch_X(I^u_{\textrm{X}},I^v_{\textrm{X}}),M_{\textrm{neg}}),
\end{equation}
where $X \in \{\textrm{ir}, \textrm{vis}\}$, $I^u_X$ is the input from the $u$-th image, and $I^v_X$ is a different randomly selected image.
$\hat{I}^u_X$ is the decomposed component obtained from the corresponding $u$-th image pair's fusion result.
The terms $M_{\textrm{pos}}$ and $M_{\textrm{neg}}$ are reference metric values computed using MATLAB for the corresponding positive and negative pairs.



The environment branch is supervised by predicting an environment-aware weighting term $ENV$ using the original visible image. 
The associated loss is:
\begin{equation}
\mathcal{L}_{\textrm{env}} = \mathcal{L}_{\textrm{mse}}\big(Branch_{\textrm{env}}(I_{\textrm{vis}}), ENV(I_{\textrm{vis}})\big), \label{eq_env_loss}
\end{equation}
where $Branch_{\textrm{env}}$ denotes the output of the environment branch, and $ENV(I_{\textrm{vis}})$ is the environment label described in Section~\ref{sec_env}.

\section{Consistency Evaluation}
Although numerous image fusion metrics have been proposed, the consistency of these metrics, namely, how well their numerical scores align with the perceived visual quality of fusion results, remains unexplored in the infrared and visible image fusion (IVIF) domain.
Prior works often address this issue only through qualitative discussions or anecdotal visual comparisons, lacking a systematic and objective evaluation of metric reliability.
To bridge this gap, we introduce a metric consistency evaluation protocol designed to quantify the agreement between metric-based rankings and independent reference standards.
Its sole aim is to provide a fair and objective lens through which to assess the reliability of various image fusion metrics.

Rather than relying on subjective human ratings, which can introduce bias and inconsistency, we utilise third-party models as reference standards.
These include no-reference image quality assessments and downstream task models (\textit{e.g.}, object detectors or semantic segmenters), which reflect real-world applicability.
For a given image fusion method $i$, let $R_i^{\textrm{Ref}}$ denote its ranking according to such a reference model.
For an arbitrary metric $M$, the corresponding ranking is denoted $R_i^M$.

We define the ranking discrepancy of metric $M$ for method $i$ as:
\begin{equation}
\Delta R_i = |R_i^M - R_i^{\textrm{Ref}}|.
\end{equation}

To account for the fact that higher-ranked methods are typically of greater interest, we propose a bi-directional weighting strategy:
\begin{equation}
W_i = \frac{1}{2}(\alpha^{R_i^M} + \beta^{R_i^{\textrm{Ref}}}),
\end{equation}
where $\alpha, \beta \in (0, 1)$ are decay factors.
This weighting mechanism ensures that methods ranked highly by the reference standard receive more attention ($\alpha^{R_i^{\textrm{Ref}}}$), and metric $M$ is more strictly penalised for errors in top-ranked methods ($\beta^{R_i^M}$).
We set $\alpha = \beta$ to treat both perspectives symmetrically.

The overall consistency score of a given metric is computed as:
\begin{equation}
MC = \exp\left(-s \cdot \sum_{i=1}^L W_i \cdot \Delta R_i\right),
\end{equation}
where $L$ is the number of evaluated methods, and $s$ is a scaling factor to control the influence of ranking error.
A higher $MC$ value indicates greater alignment between the metric and the reference.

Fusion results of 16 representative methods, from traditional approaches to advanced models, including IFCNN~\cite{zhang2020ifcnn}, U2Fusion~\cite{xu2020u2fusion}, SD-Net~\cite{zhang2021sdnet}, RFN-Nest~\cite{li2021rfn}, TarDAL~\cite{liu2022target}, YDTR~\cite{tang2022ydtr}, ReCoNet~\cite{huang2022reconet}, CDDFuse~\cite{zhao2023cddfuse}, MetaFusion~\cite{Zhao2023metafusion}, DDFM~\cite{Zhao_2023_ICCV_DDFM}, MUFusion~\cite{cheng2023mufusion}, LRRNet~\cite{li2023lrrnet}, Text-IF~\cite{yi2024textIF}, EMMA~\cite{zhao2024emma}, CoCoNet~\cite{liu2024coconet}, and FusionBooster~\cite{cheng2025fusionbooster}, are regarded as competitors to obtain the initial ranking results.

Depending on the reference model used, we instantiate two forms of this consistency score: when a deep no-reference image quality model~\cite{bosse2017deepiqa} is employed as the reference, we obtain $\textrm{MC}_{\textrm{deep}}$, which reflects perceptual quality consistency;
when a downstream task model (\textit{e.g.}, object detection) is used as the reference, the resulting $\textrm{MC}_{\textrm{ds}}$ reflects task-driven consistency.
\myrevisedcolor{
We emphasise that the goal of this protocol is not to force all fusion metrics to agree with a single criterion, but to provide an objective reliability lens under the no-GT infrared and visible image fusion setting.
No-reference IQA models offer a task-agnostic proxy for perceptual plausibility, whereas downstream task performance reflects task relevance in practical deployments.
Either reference alone can be incomplete: IQA may overlook task-critical cues, while a specific task model may bias the evaluation towards one applicatio.
By reporting both metrics, we obtain complementary perspectives and can identify metrics whose rankings are consistently supported by independent references.}

\section{Experiments}


\begin{table}[tbp]
\centering
\caption{\myrevisedcolor{Comparison of runtime evaluation settings between traditional image fusion metrics and the proposed EvaNet. All experiments are conducted under an identical workstation.}}
\resizebox{1\linewidth}{!}{
\begin{tabular}{c|c|c}
\hline
Setting & Traditional Metrics & EvaNet (Ours) \\
\hline
Platform & MATLAB & PyTorch \\
CPU & Intel i7-6850K & Intel i7-6850K \\
GPU & -- & NVIDIA RTX 3090 \\
Batch size & 1 & 1 \\
Image resolution & 640$\times$512 & 640$\times$512 \\
I/O time included & No & No \\
Execution manner & Sequential metric calls & Single forward pass \\
Metrics per run & Single metric & 8 metrics simultaneously \\
Dataset & LLVIP-test & LLVIP-test \\
\hline
\end{tabular}
}
\label{tab:runtime_comparison}
\end{table}

\subsection{Experimental Setting}
During inference, EvaNet takes the fused image as input and directly outputs the required $N$ image fusion assessments.
\myrevisedcolor{
The information probes used for decomposing the fused image are directly adopted from FusionBooster~\cite{cheng2025fusionbooster} and are pre-trained on the LLVIP training set.
These probes are fixed during both training and inference of EvaNet.
For feature extraction in the modality and environment branches, we employ a VGG backbone pre-trained on ImageNet, which serves as a generic perceptual feature extractor.}
As deep investigations into evaluation protocols for infrared and visible image fusion (IVIF) remain limited, we focus our analysis on the consistency of evaluation metrics across three variants: (1) the original formulation of each metric, (2) its improved version with textual guidance as proposed in TextFusion~\cite{cheng2023textfusion}, and (3) our EvaNet-based predictions.
EvaNet is trained using the Adam optimiser with a learning rate of $1\times10^{-3}$ and a batch size of 8.
The decay factors for consistency evaluation, $\alpha$ and $\beta$, are both set to 0.9.
\myrevisedcolor{All runtime evaluations are conducted on the same workstation under identical hardware conditions.
Due to the fundamentally different computational paradigms, classical metrics relying on CPU-based, sequential signal processing and EvaNet operating as a unified, GPU-accelerated forward-pass predictor, perfectly matched CPU--GPU comparisons are inherently infeasible.
We therefore report all runtime results transparently and summarise the controlled evaluation settings in Table.~\ref{tab:runtime_comparison}.}



\begin{table*}[tbp]
  \centering
  \caption{\myrevisedcolor{The speed (in seconds) of generating the assessment scores using eight different metrics and our EvaNet, measured across part (250 pairs) and all fused images (3463 pairs) from the LLVIP dataset. (\textbf{Bold}: Best performance)}}
  
  \resizebox{1\linewidth}{!}{ 
    \begin{tabular}{ccccccccccc}
    \hline
    Dataset & PSNR  & CC    & Qabf  & VIF   & SSIM  & $\textrm{FMI}_{\textrm{pixel}}$ & $\textrm{FMI}_{\textrm{dct}}$ & $\textrm{FMI}_{\textrm{w}}$ & Sum (8 metrics)   & \cellcolor[rgb]{ .749,  .749,  .749}EvaNet (8 metrics) \\
    \hline
    LLVIP-test (part) & 0.2s  & 1.0s  & 27.2s & 330.4s & 9.6s  & 516.1s & 2770.4s & 4039.4s & 7694.3s ($>$2 hours) & \cellcolor[rgb]{ .749,  .749,  .749}\textbf{1.6s} ($<$1 min) \\
    LLVIP-test (all) & 2.9s  & 13.5s & 341.3s & 4560.6s & 111.5s & 8121.4s & 38571.3s & 51094.5s & 102817.0s ($>$24 hours) & \cellcolor[rgb]{ .749,  .749,  .749}\textbf{10.8s} ($<$1 min) \\
    \hline
    \end{tabular}%
}    
  \label{table_inference_time}%
\end{table*}%

\begin{table}[tbp]
  \centering
  \caption{A summary of the computational characteristics of existing image fusion evaluation methods when delivering $N$ assessment metrics.}
  
  \resizebox{1\linewidth}{!}{ 
    \begin{tabular}{ccccc}
    \hline
    Method & Setting & Manner & Efficiency & Model Size \\
    \hline
    Vanilla (N metrics) & Traditional & N-Exe. & 30 s / img & - \\
    TextAttn (N metrics) & Traditional & 2N-Exe. & 60 s / img & 1552.67 MB \\
    \rowcolor[rgb]{ .749,  .749,  .749} EvaNet (N metrics) & Leaning-based & \textbf{1 Exe.} & \textbf{6 ms / img} & \textbf{1.15 MB} \\
    \hline
    \end{tabular}%
    }
  \label{table_metrics_statistics}%
\end{table}%

\subsection{A Comparison of the Speed of Evaluation of Different Image Fusion Metrics}
To demonstrate the efficiency of our EvaNet, we benchmark the inference speed of existing image fusion metrics.
As shown in Table~\ref{table_inference_time} and Table~\ref{table_metrics_statistics}, the traditional evaluation pipelines require considerable time to deliver a comprehensive assessment across multiple metrics, often due to their reliance on individual, sequential executions of handcrafted algorithms.

While methods such as TextFusion introduce textual guidance to enhance metric interpretability, they incur additional computational costs arising from the use of large-scale pre-trained vision-language models.
The need to compute the association maps and cross-modal embeddings significantly undermines their runtime efficiency, despite their conceptual advantages.

In contrast, our EvaNet eliminates such computational bottlenecks. By adopting a compact and efficient architecture, occupying merely \textasciitilde 1.15 MB in size, EvaNet directly predicts multiple fusion metrics in a single forward pass.
This streamlined and perceptually aware design enables EvaNet to deliver all 8 evaluation metrics with exceptional speed, achieving comprehensive inference within just 1.6 seconds, compared to over 7000 seconds using traditional methods.
\myrevisedcolor{These advantage become even more significant when evaluating the fusion performance based on the whole LLVIP test set (see the second row of Table~\ref{table_inference_time}).
It still cost EvaNet less than one minute to deliver all 8 metrics, while a single evaluation on traditional metric $\textrm{FMI}_{\textrm{d}}$ can take more than two hours to obtain the final result.
Note that, the efficiency gain does not solely stem from GPU versus CPU execution, but more fundamentally from the paradigm shift from classical signal-processing-based metric computation to a learning-based, unified prediction framework.
} 

\begin{figure}[t]
\centering
\includegraphics[width=1\linewidth]{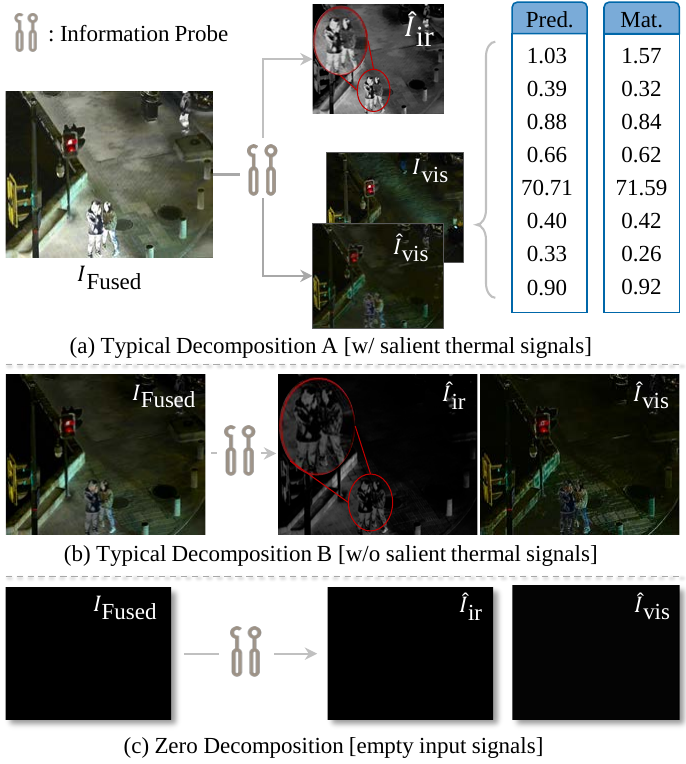}
\caption{
\myrevisedcolor{
Typical and zero-input decomposition results produced by the Information Probe.
(a) and (b) show successful disentanglement of modality-specific content from two representative fused images with different degrees of thermal information preservation.
(c) presents the zero-decomposition case, where an all-zero input yields two clean blank outputs without artefacts, verifying the robustness of the probe and its non-reliance on memorised patterns.
At the right side of (a), we present intuitive prediction results of our EvaNet for 8 image fusion assessments, \textit{i.e.}, VIF, Qabf, SSIM, CC, PSNR, $\textrm{FMI}_{\textrm{dct}}$, $\textrm{FMI}_{\textrm{wave}}$, and $\textrm{FMI}_{\textrm{pixel}}$, for the decomposed image and the source input.}
}
\label{figure_qualitative_decompisition}
\end{figure}

\subsection{The Decomposition Results of the Information Probe}
In this section, we present the decomposition capability of the Information Probe proposed in FusionBooster~\cite{cheng2025fusionbooster}.
As shown in Fig.~\ref{figure_qualitative_decompisition} (a), the probe successfully separates a fused image into two outputs that retain modality-specific characteristics.
For example, one output emphasises thermal regions, while the other preserves detailed texture information, reflecting a meaningful and interpretable disentanglement process.

\myrevisedcolor{Importantly, the decomposition behaviour is consistent with the fusion characteristics of different methods.
For instance, compared with TarDAL (Fig.~\ref{figure_qualitative_decompisition} (a)), the infrared component decomposed from U2Fusion (Fig.~\ref{figure_qualitative_decompisition} (b)) exhibits less salient thermal responses, which aligns well with the known tendency of U2Fusion to suppress strong thermal signals in the fused output.
Meanwhile, the corresponding visible component faithfully reflects the texture retention of the fused image.
These observations indicate that the decomposed components are derived from the fused image content itself, rather than being artificially synthesised.}

On the right side of Fig.~\ref{figure_qualitative_decompisition} (a), we report EvaNet’s predictions for eight widely used image fusion metrics based on the original source images $I_X$ and the decomposed components $\hat{I}_X$.
The predicted values closely follow the ground truth values obtained using standard MATLAB implementations. Although minor numerical deviations may exist, the use of a shared evaluation network across all samples ensures that the metric-based rankings remain reliable and comparable across different fusion methods.

To validate the structural integrity of the decomposition further, we test the Information Probe with an all-zero input image (Fig.~\ref{figure_qualitative_decompisition} (c)).
As visualised in the bottom section of the figure, the model produces two blank outputs without introducing artificial details.
This result further confirms that the module performs genuine decomposition based on input content, rather than relying on memorised patterns or generating artefacts from noise.

\begin{table*}[tbp]
  \centering
  \caption{\myrevisedcolor{Ablation experiments of the different components of the proposed EvaNet when applied to 8 different image fusion metrics. ``$\Delta$ vs. (a) " indicates the average metric performance change compared to the setting a.}}
  
  \resizebox{0.8\linewidth}{!}{ 
    \begin{tabular}{cccccccccc}
    \hline
    Setting & $\Delta$ vs. (a) & $\textrm{VIF}^*$ & $\textrm{Qabf}^*$ & $\textrm{SSIM}^*$ & $\textrm{CC}^*$ & $\textrm{PSNR}^*$ & $\textrm{FMI}_{\textrm{d}}^*$ & $\textrm{FMI}_{\textrm{w}}^*$ & $\textrm{FMI}_{\textrm{p}}^*$ \\
    \hline
    (a) Traditional & - & \textbf{0.728 } & 0.685  & 0.650  & 0.635  & 0.611  & 0.619  & 0.603  & 0.655  \\
    (b) Dec. only & 2.27\% & 0.673  & 0.620  & 0.623  & 0.664  & 0.727  & \textbf{0.713 } & 0.647  & 0.614  \\
    (c) Con. only & -0.33\% & 0.725  & 0.642  & 0.648  & 0.615  & 0.619  & 0.633  & 0.618  & 0.664  \\
    (d) Env. only & -1.37\% & 0.655  & 0.692  & 0.657  & 0.658  & 0.622  & 0.564  & 0.617  & 0.643  \\
    (e) w/o Env. & 7.47\% & 0.710  & 0.682  & 0.717  & 0.724  & 0.716  & 0.691  & 0.677  & 0.639  \\
    (f) w/o Con. & -0.80\% & 0.587  & 0.605  & 0.683  & 0.665  & 0.730  & 0.625  & 0.608  & 0.614  \\
    (g) w/o Dec. & 2.16\% & 0.669  & 0.674  & 0.650  & 0.719  & 0.648  & 0.614  & 0.660  & 0.651  \\
    \rowcolor[rgb]{ .851,  .851,  .851} EvaNet (Ours) & 8.70\% & 0.724  & \textbf{0.717 } & \textbf{0.726 } & \textbf{0.724 } & \textbf{0.737 } & 0.640  & \textbf{0.689 } & \textbf{0.664 } \\
    \hline
    \end{tabular}%
}
  \label{table_quantitative_ablation}%
\end{table*}%

\begin{figure}[t]
\centering
\includegraphics[width=1\linewidth]{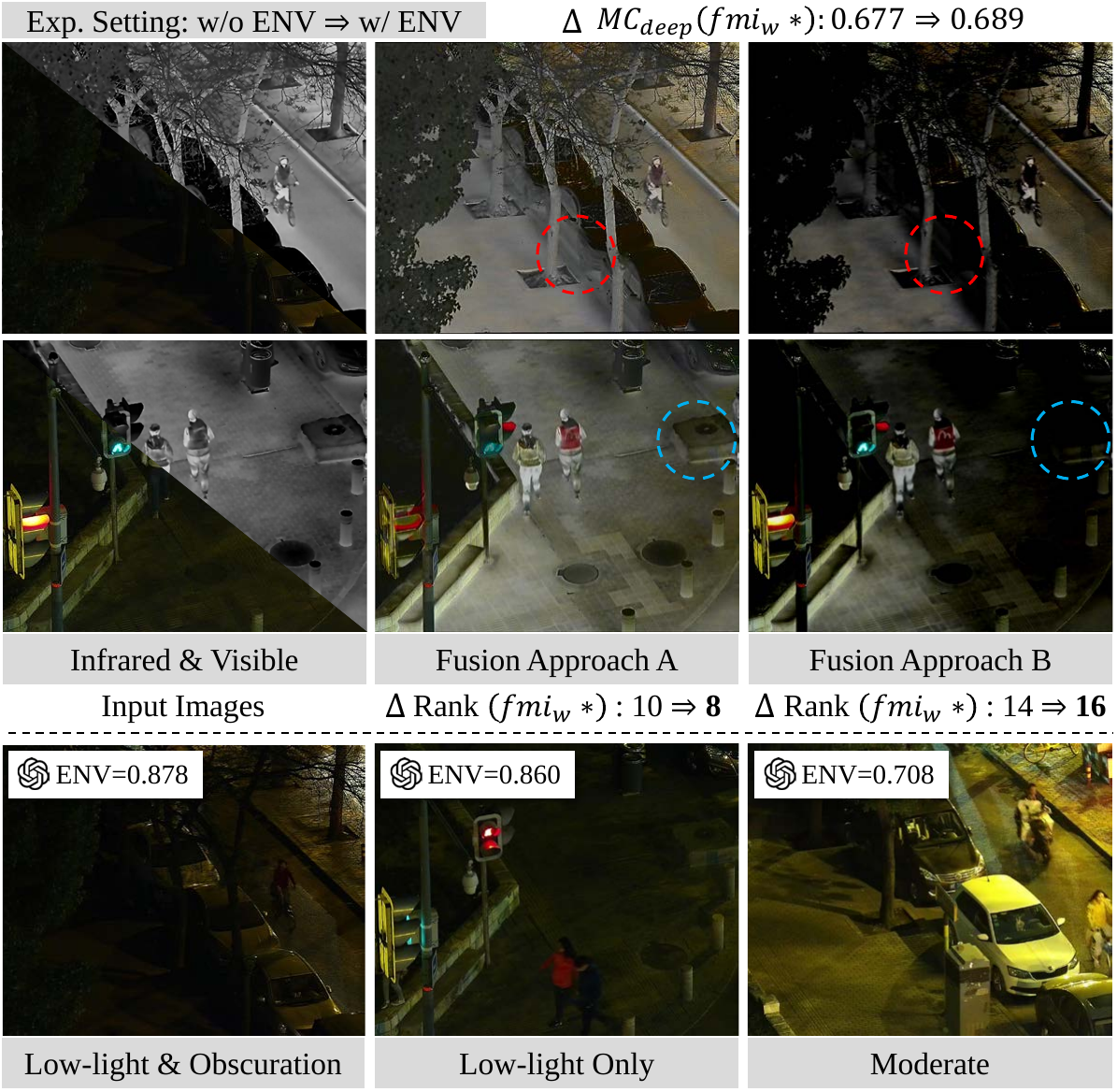}
\caption{
The fusion results of two representative approaches A (SDNet) and B (ReCoNet). Below the fused images, we present the ablation results, highlighting the impact of the environment perception branch within our EvaNet. 
``X→Y" denotes the change in the corresponding value after enabling environment-aware evaluation.
\myrevisedcolor{The proposed environment-aware design can effectively encourage and penalise results with or without an appropriate illumination condition.}
The environmental condition predicted by the perception branch is shown in each source image.
}
\label{figure_qualitative_ablation}
\end{figure}

\subsection{Ablation Studies of the Different Components}
In this section, we present both intuitive and quantitative results to assess the contribution of various components within the proposed EvaNet architecture.
Table~\ref{table_quantitative_ablation} reports the effect of removing the environment perception branch (setting (e)).
In this configuration, the consistency across multiple image fusion metrics shows a noticeable decline.
This observation suggests that the incorporation of the environmental condition assessment contributes positively and consistently to the reliability of the evaluation.

Fig.~\ref{figure_qualitative_ablation} provides further visual evidence.
When the environmental condition such as low-light ( illustrated in the rightmost example ) is considered, the model is able to penalise the relevant modality accordingly.
On the contrary, when the scene illumination is satisfactory, our environment-aware EvaNet will reflect that in assigning a favourable weighting to the visible light modality.
The lower section of the figure displays the predictions from the environment perception branch.
In cases where the scene contains compounded challenges, such as poor lighting and visual occlusion caused by trees, our perceptual scene assessment branch assigns the highest penalties, effectively avoiding an over-reliance on the less informative visible features.
Conversely, under moderately illuminated conditions, the perceived difficulty of the environment is lower, leading to a correspondingly reduced weighting. 

Additionally, we examine the role of contrastive learning. When this strategy is omitted during training (setting (f) of Table~\ref{table_quantitative_ablation}), the consistency of fusion metric predictions drops significantly.
This highlights the crucial importance of contrastive learning in shaping a more discriminative and reliable evaluation model.
Finally, setting (g) explores the impact of removing the decomposition module.
Without explicitly disentangling the fused image into its constituent modalities, EvaNet loses its capacity to perform fine-grained evaluation.
As a result, its consistency does not reach the level achieved by versions that retain this decomposition process, \textit{e.g.}, the setting (b).

\begin{figure}[t]
\centering
\includegraphics[width=1\linewidth]{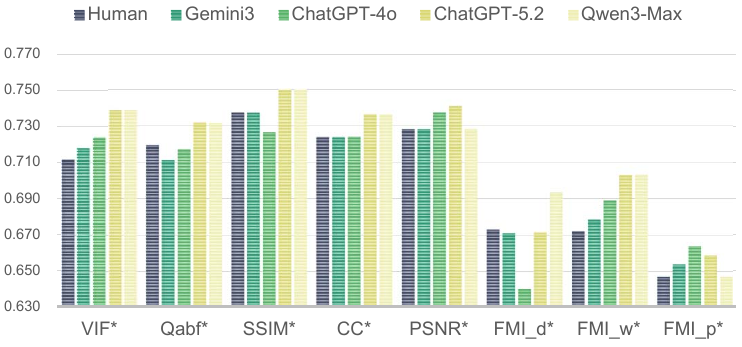}
\caption{
\myrevisedcolor{
Metric consistency scores obtained using environment labels generated by different large language models and a human-labeled baseline.
More advanced LLMs yield higher overall consistency, while all LLM-based supervisions outperform human annotation.
}
}
\label{figure_llms}
\end{figure}

\myrevisedcolor{
\subsection{Sensitivity Analysis of LLM-Based Environment Supervision}
\label{sec_llm_sensitivity}
The environment branch of EvaNet is supervised by illumination and obscuration attributes generated by a large language model (LLM).
To evaluate the sensitivity of this supervision, we compare multiple LLMs and a human-labeled baseline.
Specifically, we regenerate environment scores using Gemini~3~\cite{google_gemini3}, ChatGPT-4o, ChatGPT-5.2~\cite{OpenAI2025ChatGPT-4o}, and Qwen3-Max~\cite{alibaba_qwen3} under identical prompt templates and normalisation settings (Section~\ref{sec_env}), and recompute the corresponding metric rankings and consistency scores $\textrm{MC}_{\textrm{deep}}$.
To provide a non-LLM reference, we additionally invite a volunteer with basic knowledge of image fusion to manually annotate illumination and obscuration on a 0--5 scale, from which consistency scores are derived using the same pipeline.}

\myrevisedcolor{As summarised in Fig.~\ref{figure_llms}, different LLMs yield stable ranking behaviour with varying absolute consistency.
More advanced models, such as ChatGPT-5.2 and Qwen3-Max, consistently achieve higher consistency than ChatGPT-4o, while all LLM-based supervisions outperform the human-labeled baseline, demonstrating the stability and scalability of LLM-based environment supervision.}

\begin{table*}[tbp]
  \centering
  \caption{The quantitative results of $\textrm{MC}_{\textrm{deep}}$ on different image fusion datasets. (\textbf{Bold}: Better performance compared to the vanilla metrics)}
  
  \resizebox{0.85\linewidth}{!}{ 
    \begin{tabular}{cccccccccc}
    \hline
    Dataset & \multicolumn{3}{c}{LLVIP Dataset} & \multicolumn{3}{c}{RoadScene Dataset} & \multicolumn{3}{c}{TNO Dataset} \\
    \hline
    Metric & Vanilla  & +TextAttn & \cellcolor[rgb]{ .749,  .749,  .749}+EvaNet & Vanilla  & +TextAttn & \cellcolor[rgb]{ .749,  .749,  .749}+EvaNet & Vanilla  & +TextAttn & \cellcolor[rgb]{ .749,  .749,  .749}+EvaNet \\
    \hline
    $\textrm{MC}_{\textrm{deep}}$(VIF) & 0.634  & 0.659  & \cellcolor[rgb]{ .749,  .749,  .749}\textbf{0.724 } & 0.753  & 0.736  & \cellcolor[rgb]{ .749,  .749,  .749}0.629  & 0.621  & 0.622  & \cellcolor[rgb]{ .749,  .749,  .749}\textbf{0.638 } \\
    $\textrm{MC}_{\textrm{deep}}$(Qabf) & 0.611  & 0.607  & \cellcolor[rgb]{ .749,  .749,  .749}\textbf{0.717 } & 0.542  & 0.582  & \cellcolor[rgb]{ .749,  .749,  .749}\textbf{0.709 } & 0.609  & 0.620  & \cellcolor[rgb]{ .749,  .749,  .749}\textbf{0.644 } \\
    $\textrm{MC}_{\textrm{deep}}$(SSIM) & 0.649  & 0.560  & \cellcolor[rgb]{ .749,  .749,  .749}\textbf{0.726 } & 0.584  & 0.603  & \cellcolor[rgb]{ .749,  .749,  .749}\textbf{0.655 } & 0.603  & 0.601  & \cellcolor[rgb]{ .749,  .749,  .749}\textbf{0.645 } \\
    $\textrm{MC}_{\textrm{deep}}$(CC) & 0.696  & 0.608  & \cellcolor[rgb]{ .749,  .749,  .749}\textbf{0.724 } & 0.590  & 0.589  & \cellcolor[rgb]{ .749,  .749,  .749}\textbf{0.630 } & 0.673  & 0.614  & \cellcolor[rgb]{ .749,  .749,  .749}0.655  \\
    $\textrm{MC}_{\textrm{deep}}$(PSNR) & 0.609  & 0.559  & \cellcolor[rgb]{ .749,  .749,  .749}\textbf{0.737 } & 0.601  & 0.612  & \cellcolor[rgb]{ .749,  .749,  .749}\textbf{0.633 } & 0.667  & 0.663  & \cellcolor[rgb]{ .749,  .749,  .749}0.645  \\
    $\textrm{MC}_{\textrm{deep}} (\textrm{FMI}_\textrm{d})$ & 0.713  & 0.637  & \cellcolor[rgb]{ .749,  .749,  .749}0.640  & 0.588  & 0.576  & \cellcolor[rgb]{ .749,  .749,  .749}\textbf{0.764 } & 0.565  & 0.607  & \cellcolor[rgb]{ .749,  .749,  .749}\textbf{0.671 } \\
    $\textrm{MC}_{\textrm{deep}}(\textrm{FMI}_\textrm{w})$ & 0.672  & 0.677  & \cellcolor[rgb]{ .749,  .749,  .749}\textbf{0.689 } & 0.561  & 0.614  & \cellcolor[rgb]{ .749,  .749,  .749}\textbf{0.732 } & 0.544  & 0.588  & \cellcolor[rgb]{ .749,  .749,  .749}\textbf{0.606 } \\
    $\textrm{MC}_{\textrm{deep}}(\textrm{FMI}_\textrm{p})$ & 0.611  & 0.672  & \cellcolor[rgb]{ .749,  .749,  .749}\textbf{0.664 } & 0.621  & 0.634  & \cellcolor[rgb]{ .749,  .749,  .749}\textbf{0.712 } & 0.624  & 0.646  & \cellcolor[rgb]{ .749,  .749,  .749}0.566  \\
    \hline
    \end{tabular}%
    }
\label{table_numerical_MC_visual_deepIQA} 
\end{table*}%

\begin{figure}[t]
\centering
\includegraphics[width=1\linewidth]{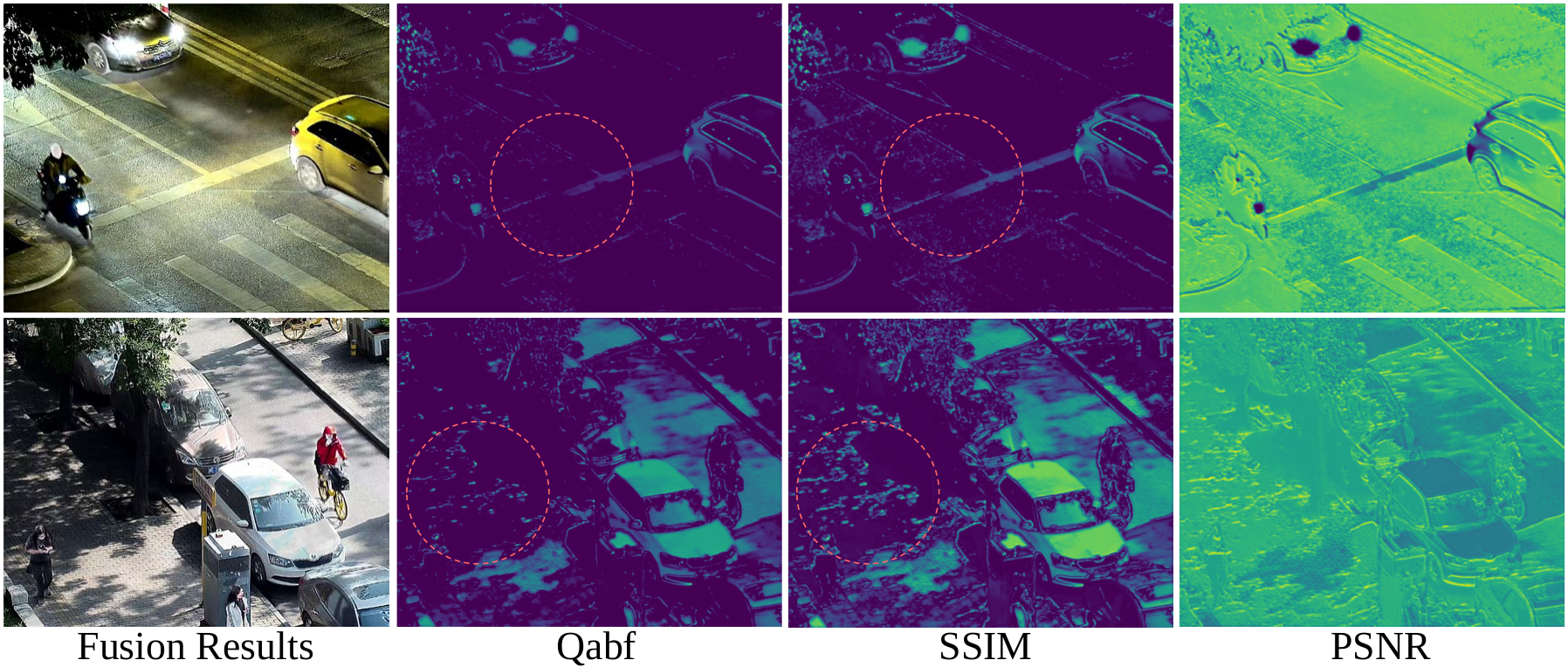}
\caption{
\myrevisedcolor{
Metric-wise attribution visualisation of ``Conv3" for two fusion results on the LLVIP dataset.
Brighter regions indicate higher contribution to the predicted metric. Different metrics attend to distinct image regions in accordance with their underlying definitions.}
}
\label{figure_qualitative_attribution_maps}
\end{figure}
\myrevisedcolor{
\subsection{Metric-wise Attribution Analysis}
To investigate the interpretability of the proposed EvaNet, we conduct an attribution-based visualisation analysis of ``Conv3" of our EvaNet on different predicted similarity metrics.
We select two fusion results during the daytime and night produced by CocoNet on the LLVIP dataset.}

\myrevisedcolor{As shown in Fig.~\ref{figure_qualitative_attribution_maps}, the attribution results reveal clear and interpretable differences among the considered metrics.
For Qabf, which is designed to evaluate whether salient information is effectively transferred from the source images to the fused result, the attribution maps predominantly highlight strong edges and salient boundaries. 
Similarly, SSIM, as a structure-oriented metric, exhibits broader spatial responses. 
In contrast, PSNR shows a notably different behaviour. The attribution maps for PSNR tend to focus on globally bright regions in the scene, such as pedestrians and vehicles.
These results demonstrate that the proposed network not only predicts multiple quality metrics accurately but also learns metric-specific attention patterns that are consistent with the underlying definitions of each metric. This analysis provides additional evidence for the interpretability and reliability of the proposed evaluation framework.}


\begin{figure}[t]
\centering
\includegraphics[width=1\linewidth]{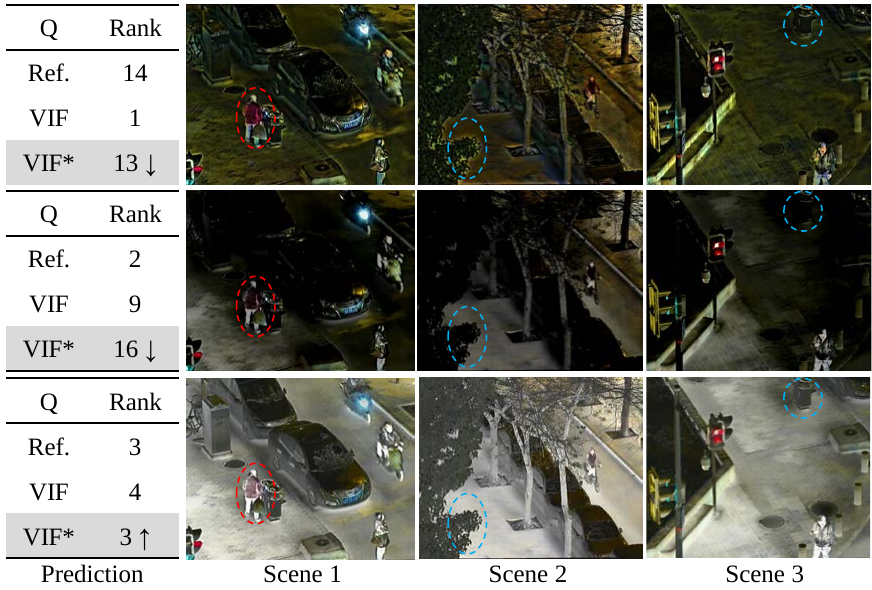}
\caption{
The qualitative results produced by the EvaNet when predicting representative fusion results on the LLVIP dataset.
From top to bottom, MetaFusion~\cite{Zhao2023metafusion}, ReCoNet~\cite{huang2022reconet}, and FusionBooster~\cite{cheng2025fusionbooster}.
}
\label{figure_qualitative_predict_llvip}
\end{figure}

\begin{figure}[t]
\centering
\includegraphics[width=1\linewidth]{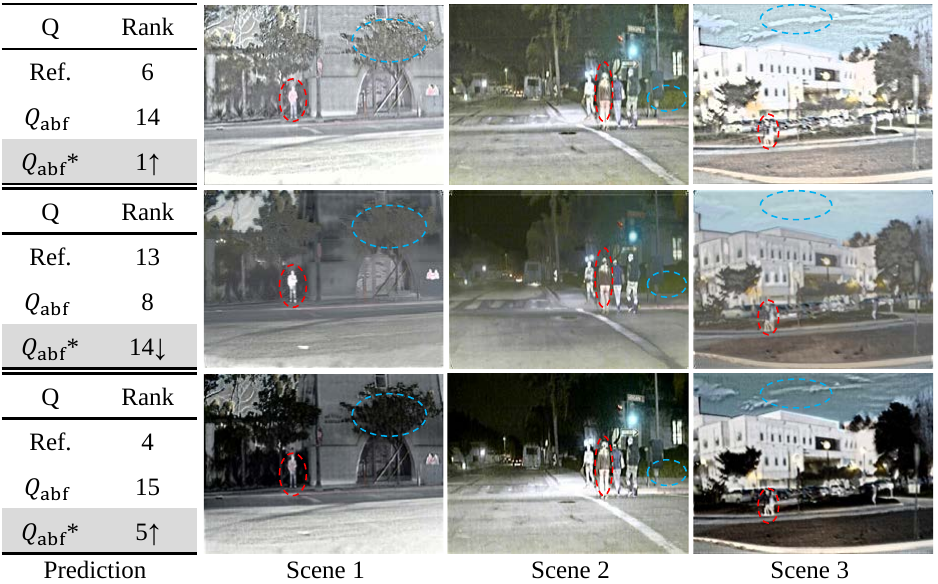}
\caption{
The qualitative results produced by the EvaNet when predicting the representative fusion results on the RoadScene dataset.
From top to bottom, MetaFusion, TarDAL~\cite{liu2022target}, and CoCoNet~\cite{liu2024coconet}.
}
\label{figure_qualitative_predict_roadscene}
\end{figure}

\begin{figure}[t]
\centering
\includegraphics[width=1\linewidth]{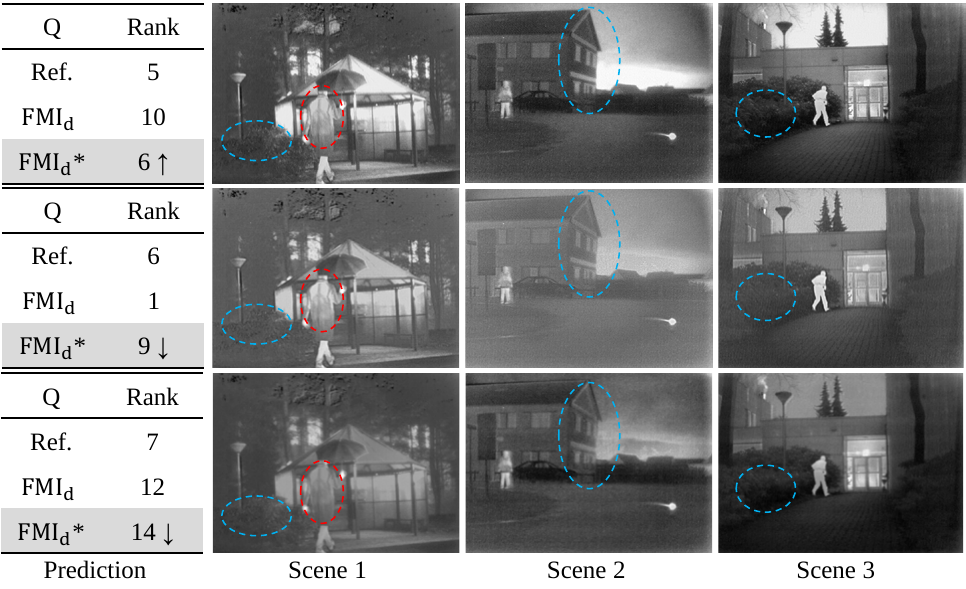}
\caption{
The qualitative results produced by the EvaNet when predicting representative fusion results on the TNO dataset.
From top to bottom, EMMA~\cite{zhao2024emma}, IFCNN~\cite{zhang2020ifcnn}, and DDFM~\cite{Zhao_2023_ICCV_DDFM}.
}
\label{figure_qualitative_predict_tno}
\end{figure}

\subsection{Metric Consistency from a Visual Perspective}

In this section, we present the consistency results of various metrics, using the widely adopted DeepIQA~\cite{bosse2017deepiqa} as a reference benchmark, \textit{i.e.}, $\textrm{MC}_{\textrm{deep}}$.
As shown in Table~\ref{table_numerical_MC_visual_deepIQA}, several traditional metrics, such as $Q_{\textrm{abf}}$, exhibit relatively low consistency when compared with other image fusion assessments.
However, once integrated with our EvaNet framework, these vanilla metrics show clear improvements in consistency.

Although the TextAttn approach incorporates additional textual guidance to enhance the metrics performance, it fails to improve the reliability of the metric outputs consistently.
In fact, it often leads to a decline in metric consistency across different evaluations.
This degradation may stem from the fact that TextAttn only emphasises text-relevant regions within the fused image, while relying heavily on the conventional methods of assessment, as outlined in Table~\ref{table_metrics_statistics}.

In contrast, EvaNet improves the consistency of most image fusion metrics significantly, while also offering substantial gains in computational efficiency.
This improvement can be attributed to its general architectural design, which incorporates both the perceptual assessment of the environmental conditions and the fine-grained evaluation of the modality-specific content.
These components allow the model to account for the complexities of different conditions better.

To support our quantitative findings, we provide qualitative examples in Fig.~\ref{figure_qualitative_predict_llvip}, Fig.~\ref{figure_qualitative_predict_roadscene}, and Fig.~\ref{figure_qualitative_predict_tno}, focusing on three metrics: VIF, Qabf, and $\textrm{FMI}_{\textrm{dct}}$.
In these figures, we display the fusion results from three representative methods (columns two to four), alongside the metric rankings (first column).
For instance, MetaFusion (top row of Fig.~\ref{figure_qualitative_predict_llvip}) introduces evident artefacts into the fused image, yet still ranks highest under the vanilla VIF metric.
This inconsistency is addressed by our EvaNet, which suppresses MetaFusion’s ranking from 1 to 13, aligning more closely with the human perceptual reference.

Conversely, when evaluating the methods that preserve significant gradient information contained in source images, EvaNet tends to reward them accordingly.
As shown in the first row of Fig.~\ref{figure_qualitative_predict_roadscene}, the texture-rich output of this method is assigned a higher ranking compared with others, accurately reflecting its visual characteristic.

A similar conclusion can be drawn from the qualitative results in Fig.~\ref{figure_qualitative_predict_tno}, which show the assessment of $\textrm{FMI}_{\textrm{dct}}$.
Here, EMMA benefits from a notable ranking improvement, rising from tenth to sixth place, due to its balanced preservation of information from both input modalities.
In contrast, methods such as IFCNN, which produce suboptimal results, are correctly downgraded within our evaluation framework.

\begin{figure}[t]
\centering
\includegraphics[width=1\linewidth]{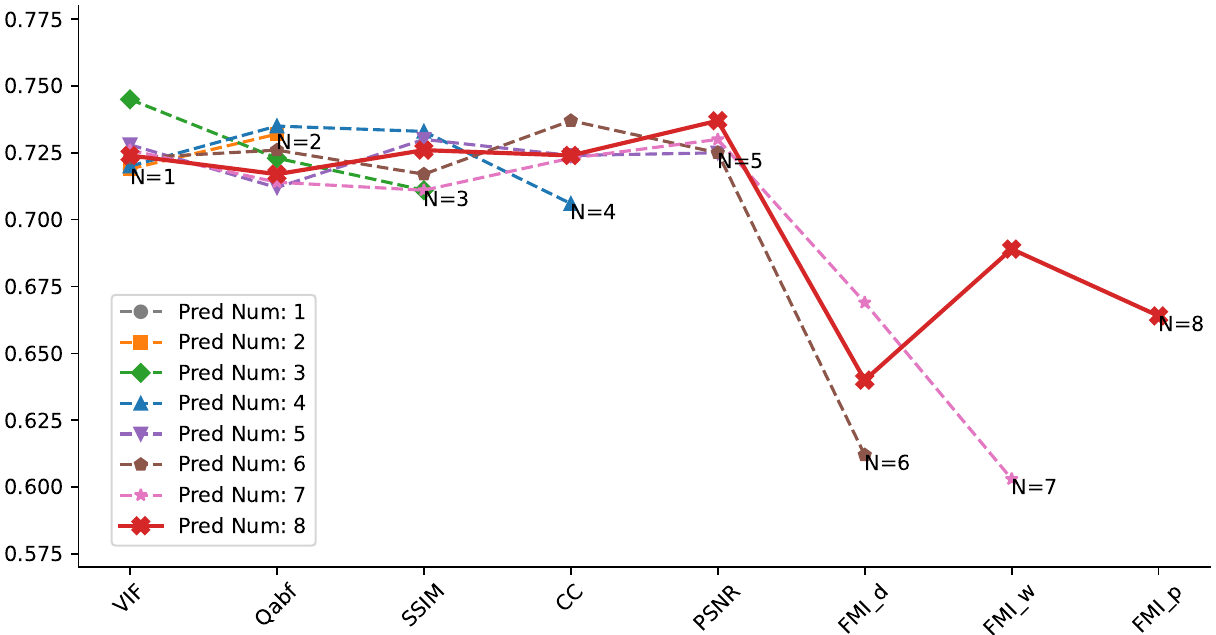}
\caption{
The quantitative results of the ablation study of the number of predicted metrics, evaluated using $\textrm{MC}_{\textrm{deep}}$.
For the first five independent metrics, increasing the number of simultaneous predictions does not significantly compromise consistency. 
This suggests that EvaNet maintains reliable performance even with a more comprehensive output configuration.
}
\label{figure_quantitative_ablation_number}
\end{figure}

\subsection{Ablation Experiments Studying the Number of Prediction Metrics}
Different image quality assessments often rely on similar underlying computations, such as gradient-based maps or mutual information.
Therefore, jointly predicting multiple metrics may enhance the overall reliability and internal consistency of the evaluation process.
In EvaNet, image fusion assessments are generated directly from the two prediction heads.
To examine whether predicting more metrics simultaneously introduces interference or, conversely, promotes consistency, we conduct a controlled ablation study by varying the number of predicted outputs.

Fig.~\ref{figure_quantitative_ablation_number} presents the quantitative consistency results measured using $\textrm{MC}_{\textrm{deep}}$ across different prediction configurations.
For the first five metrics, namely VIF, Qabf, SSIM, CC, and PSNR, we observe that increasing the number of predicted outputs does not significantly degrade metric consistency.
This indicates that these relatively independent assessments are robust to performing a joint prediction.

In contrast, for metrics related to mutual information of feature distributions, we observe a different pattern.
When only $\textrm{FMI}_{\textrm{dct}}$ is predicted (N = 6), its consistency score is suboptimal.
However, once additional related metrics such as $\textrm{FMI}_{\textrm{wave}}$ (N = 7) and $\textrm{FMI}_{\textrm{pixel}}$ (N = 8) are included, the consistency of $\textrm{FMI}_{\textrm{dct}}$ notably improves.
A similar trend is observed for $\textrm{FMI}_{\textrm{wave}}$
These findings suggest that metrics sharing common properties can mutually reinforce each other when predicted in parallel.

\begin{figure}[t]
\centering
\includegraphics[width=0.8\linewidth]{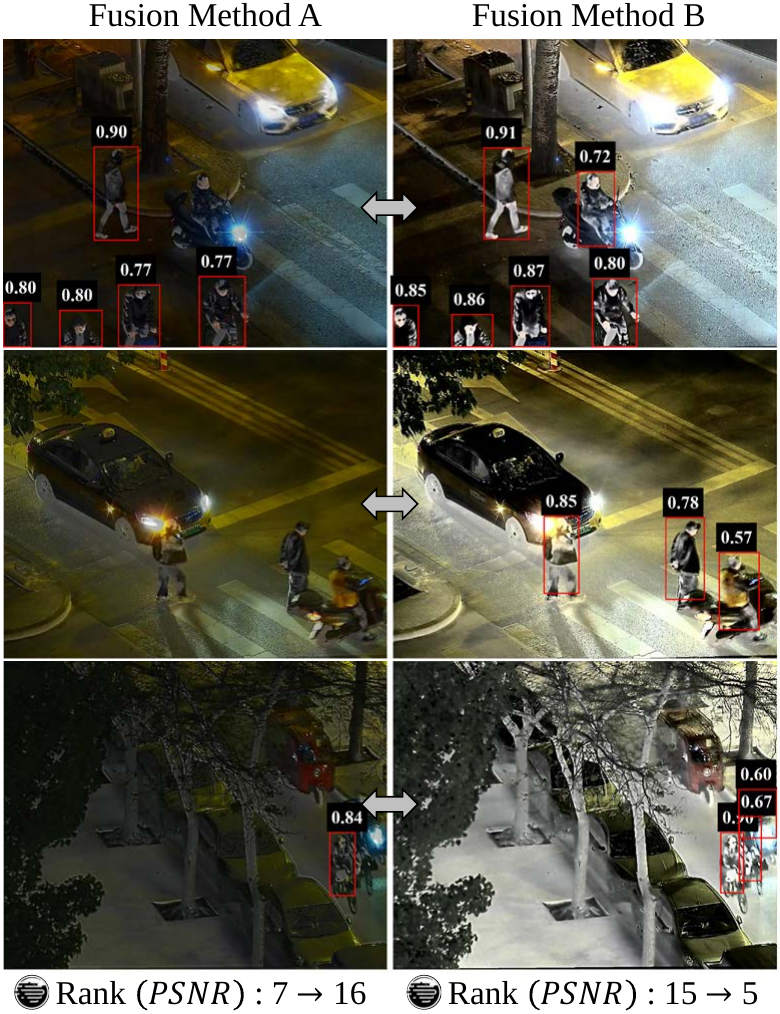}
\caption{
Detection results visualisation of two representative image fusion methods (LRRNet~\cite{li2023lrrnet} and CoCoNet~\cite{liu2024coconet}) and the corresponding ranking difference relating to PSNR with or without EvaNet.
}
\label{figure_qualitative_detection}
\end{figure}

\begin{figure}[t]
\centering
\includegraphics[width=1\linewidth]{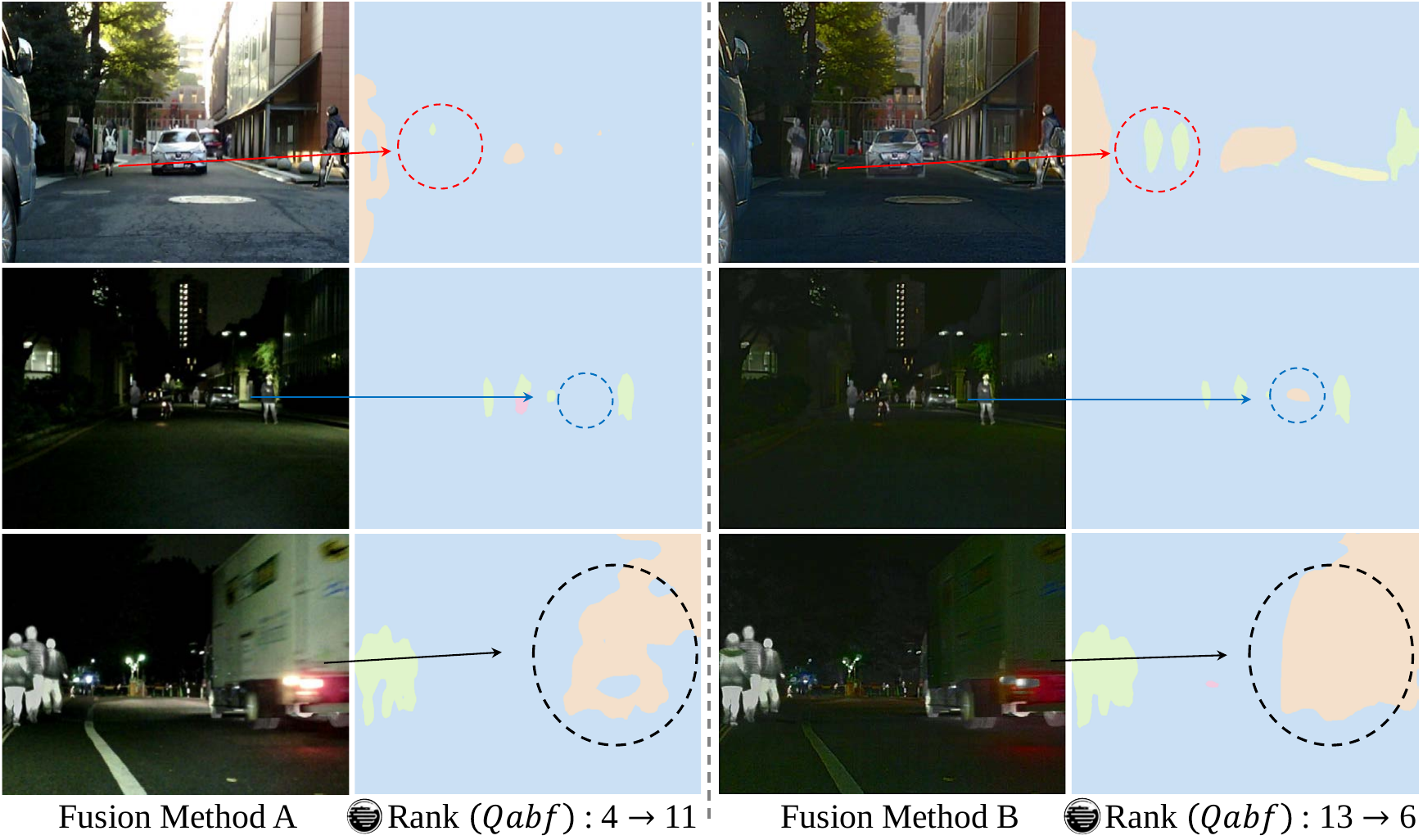}
\caption{
Segmentation results visualisation of two representative image fusion methods (EMMA~\cite{zhao2024emma} and SDNet~\cite{zhang2021sdnet}) and the corresponding ranking difference in Qabf with or without EvaNet.
}
\label{figure_qualitative_segmentation}
\end{figure}

\subsection{Metric Consistency from the Perspective of the Downstream Task Performance}
In Table~\ref{table_numerical_MC_ds_detection_segmentation}, we present the consistency results ($\textrm{MC}_{\textrm{ds}}$) obtained when using the downstream task performance as a reference, focusing on object detection~\cite{redmon2016yolo} and semantic segmentation~\cite{xu2023pidnet}.
As observed in the table, our EvaNet improves metric consistency under this stricter evaluation setting.
In the detection task, six out of eight metrics achieve higher consistency compared with their vanilla counterparts.
In the segmentation task, five out of eight metrics show similar improvements.
Notably, VIF and Qabf exhibit substantial gains in consistency within the detection setting, with increases of 0.162 (27\%) and 0.116 (19\%) respectively.
These results suggest that the corresponding metrics, when enhanced by EvaNet, align more closely with the actual performance of downstream detection tasks.

This finding supports the conclusion that our EvaNet framework offers a more task-relevant evaluation of fusion quality, as the quality of the fused image directly impacts the effectiveness of subsequent vision tasks~\cite{liu2022target}.
Consequently, researchers and practitioners may place greater emphasis on these improved metrics, particularly on the visual information fidelity (VIF) and salient gradient preservation (Qabf), when tuning fusion algorithms for downstream performance.

To further illustrate this point, we provide qualitative results for pedestrian detection in Fig.~\ref{figure_qualitative_detection}.
In low-light conditions, PSNR is traditionally expected to reflect the extent to which thermal information from the infrared source is retained in the fused result.
However, due to its pixel-level averaging, PSNR fails to capture this aspect effectively.
As shown in the figure, Fusion Method A generates fused outputs with poor thermal saliency and low detection quality (including missed detections and weak confidence scores), yet still receives a relatively high PSNR ranking (rank 7).
In contrast, Fusion Method B produces more salient pedestrian regions and achieves higher detection accuracy, but is undervalued by the PSNR metric.
EvaNet successfully corrects this discrepancy by increasing the ranking of Method B and lowering that of Method A.
However, the overall consistency score for PSNR remains relatively unchanged.
This can be attributed to the marginal difference in actual detection performance between these two methods (their mAP rankings are 14 and 13, respectively), meaning that although the metric rankings were adjusted appropriately, the downstream ground truth reference did not exhibit sufficient separation to reflect the change quantitatively.

Turning to Qabf, which is designed to capture the preservation of salient edge and gradient information from the source images, Fig.\ref{figure_qualitative_segmentation} presents a comparison between the outputs of two representative methods: EMMA\cite{zhao2024emma} and SDNet~\cite{zhang2021sdnet}.
In this example, Method B demonstrates stronger edge clarity, enabling the segmentation model to  delineate the object boundaries, such as “person” and “car”, better.
However, the original Qabf metric fails to recognise this advantage, assigning Method B a lower ranking.
EvaNet addresses this inconsistency by increasing the ranking of Method B from 13 to 6, reflecting its superior edge preservation and segmentation quality better. This highlights the ability of EvaNet to refine the behaviour of traditional metrics, aligning them more closely with perceptual and task-level relevance.

\subsection{Metric Selection across Different Datasets}

In addition to demonstrating the effectiveness of the proposed EvaNet, our consistency evaluation framework also offers insights into the relative suitability of different image fusion metrics under various conditions.
As illustrated in Table~\ref{table_numerical_MC_visual_deepIQA} and Table~\ref{table_numerical_MC_ds_detection_segmentation}, the proposed consistency scores, $\textrm{MC}_{\textrm{deep}}$ for the human visual perception and $\textrm{MC}_{\textrm{ds}}$ for the downstream task performance, reveal noticeable variations in the behaviour of individual metrics across datasets.
For instance, when the objective is to align with human perception, the enhanced version of visual information fidelity ($VIF*$) demonstrates superior consistency on the LLVIP and TNO datasets, making it a more suitable choice in perceptual studies.
Conversely, when the focus shifts towards preserving critical information for downstream tasks such as detection or segmentation, the improved version of $Q_{abf}$ ($Qabf*$) exhibits a more reliable performance.

This observation highlights the practical value of our consistency-based analysis, which enables more informed metric selection tailored to specific evaluation goals and dataset characteristics.

\begin{figure*}[t]
\centering
\includegraphics[width=0.95\linewidth]{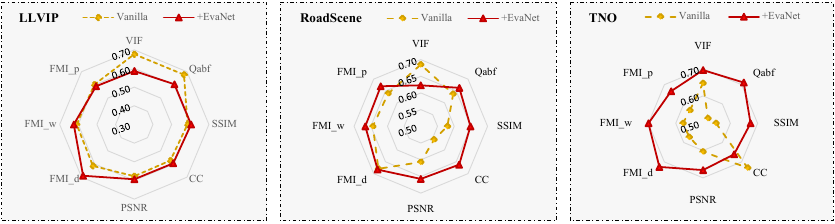}
\caption{
The quantitative results reporting metric consistency using CLIP-IQA as the reference model, \textit{i.e.}, $\textrm{MC}_{\textrm{clip}}$ values for different metrics.
}
\label{figure_clip_iqa_quantitative}
\end{figure*}

\begin{table}[tbp]
  \centering
  \caption{The quantitative results of MC-DS obtained for different downstream vision tasks. (\textbf{Bold}: Better performance compared to the vanilla metrics;'-' suggests that the metric does not support this dataset)}
  \resizebox{1\linewidth}{!}{ 
    \begin{tabular}{ccccccc}
    \hline
    \multicolumn{1}{c}{Task} & \multicolumn{3}{c}{Detection} & \multicolumn{3}{c}{Segmentation} \\
    \hline
    \multicolumn{1}{c}{Method} & Vanilla  & +TextAtt & \cellcolor[rgb]{ .749,  .749,  .749}+EvaNet & Vanilla  & +TextAtt & \cellcolor[rgb]{ .749,  .749,  .749}+EvaNet \\
    \hline
    $\textrm{MC}_{\textrm{ds}}(\textrm{VIF})$ & 0.590  & 0.575  & \cellcolor[rgb]{ .749,  .749,  .749}\textbf{0.752 } & 0.591  & -     & \cellcolor[rgb]{ .749,  .749,  .749}\textbf{0.682 } \\
    $\textrm{MC}_{\textrm{ds}}(\textrm{Qabf})$ & 0.604  & 0.615  & \cellcolor[rgb]{ .749,  .749,  .749}\textbf{0.720 } & 0.595  & -     & \cellcolor[rgb]{ .749,  .749,  .749}\textbf{0.632 } \\
    $\textrm{MC}_{\textrm{ds}}(\textrm{SSIM})$ & 0.707  & 0.607  & \cellcolor[rgb]{ .749,  .749,  .749}\textbf{0.724 } & 0.689  & -     & \cellcolor[rgb]{ .749,  .749,  .749}\textbf{0.737 } \\

    $\textrm{MC}_{\textrm{ds}}(\textrm{FMI}_\textrm{d})$ & 0.740  & 0.634  & \cellcolor[rgb]{ .749,  .749,  .749}0.654  & 0.666  & -     & \cellcolor[rgb]{ .749,  .749,  .749}0.563  \\
    $\textrm{MC}_{\textrm{ds}}(\textrm{FMI}_\textrm{w})$ & 0.642  & 0.607  & \cellcolor[rgb]{ .749,  .749,  .749}\textbf{0.658 } & 0.695  & -     & \cellcolor[rgb]{ .749,  .749,  .749}0.554  \\
    $\textrm{MC}_{\textrm{ds}}(\textrm{CC})$ & 0.717  & 0.575  & \cellcolor[rgb]{ .749,  .749,  .749}\textbf{0.746 } & 0.740  & -     & \cellcolor[rgb]{ .749,  .749,  .749}\textbf{0.775 } \\
    $\textrm{MC}_{\textrm{ds}}(\textrm{PSNR})$ & 0.775  & 0.689  & \cellcolor[rgb]{ .749,  .749,  .749}0.717  & 0.643  & -     & \cellcolor[rgb]{ .749,  .749,  .749}\textbf{0.769 } \\    
    $\textrm{MC}_{\textrm{ds}}(\textrm{FMI}_\textrm{p})$ & 0.690  & 0.724  & \cellcolor[rgb]{ .749,  .749,  .749}\textbf{0.693 } & 0.621  & -     & \cellcolor[rgb]{ .749,  .749,  .749}0.615  \\
\cline{1-4}\cline{5-7}    \end{tabular}%
}
  \label{table_numerical_MC_ds_detection_segmentation} 
\end{table}%

\subsection{The Results from the  Perspective of Subjective Visualisation}
While we have already reported the consistency performance using $\textrm{MC}_{\textrm{deep}}$, it is also important comprehensively to validate the effectiveness of our proposed approach.
With the advancement of multimodal large models, text information has increasingly been used to enhance image quality assessment.
In this section, we present the consistency results, using CLIP-IQA~\cite{wang2023clipiqa} as the reference model, replacing DeepIQA in the calculation of the consistency scores ($\textrm{MC}_{\textrm{clip}}$).

As illustrated in Fig.~\ref{figure_clip_iqa_quantitative}, EvaNet improves the metric consistency of the various baseline image fusion assessments.
Under this alternative reference framework, our method performs particularly well in enhancing the consistency of FMI-based metrics, as well as PSNR and SSIM.

It is worth noting that, when using DeepIQA as a reference, EvaNet’s performance on the TNO dataset was relatively limited.
However, the introduction of large-scale training on image–text pairs in CLIP-IQA allows it to capture the advantages offered by our evaluation method better.
For instance, the consistency score of Qabf improves from 0.5 to 0.7, confirming that our approach evaluates the ability of fusion methods to preserve the critical information coveyed by the source images more effectively.

\begin{figure}[t]
\centering
\includegraphics[width=1\linewidth]{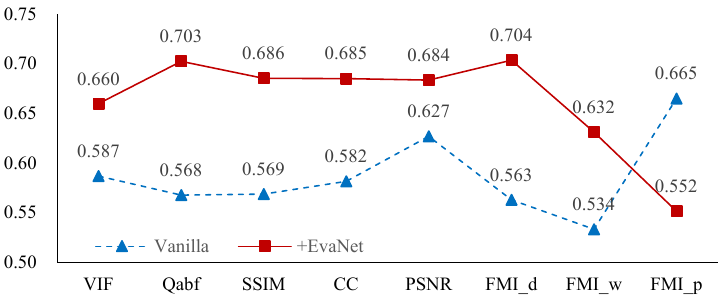}
\caption{
The results of comparison of the proposed EvaNet and the original image quality assessments on the MSRS dataset.
}
\label{figure_quantitative_msrs}
\end{figure}

\begin{figure*}[t]
\centering
\includegraphics[width=0.85\linewidth]{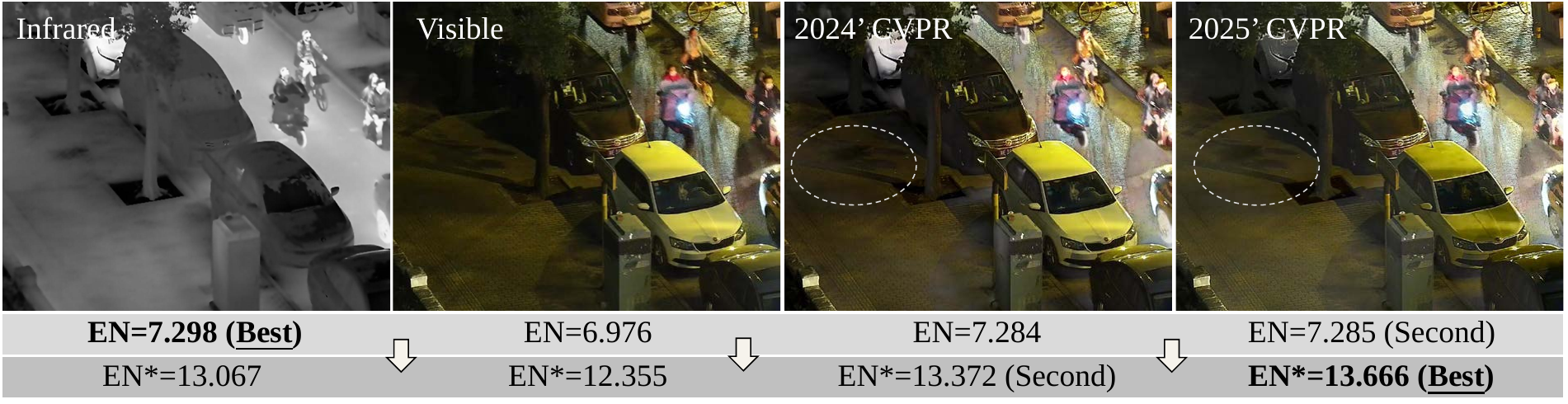}
\caption{
A comparison of the EN values of the source images and the two state-of-the-art (Text-IF~\cite{yi2024textIF} and GIFNet~\cite{cheng2025gifnet}) image fusion results.
Without an appropriate adaptation, the vanilla version of EN indicates that simply outputting the infrared modality is better than fusing two modalities together.
}
\label{figure_qualitative_noreference}
\end{figure*}

\begin{figure}[t]
\centering
\includegraphics[width=1\linewidth]{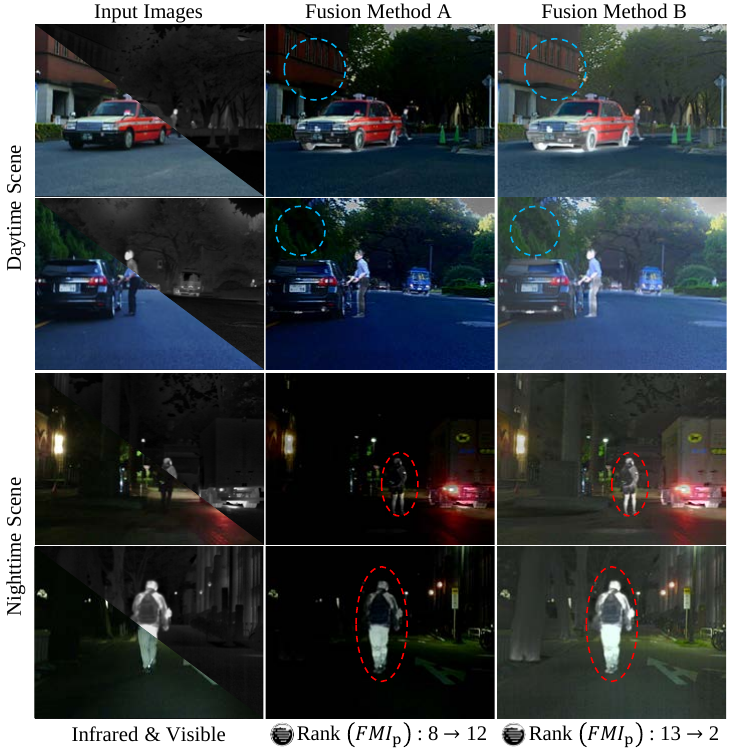}
\caption{
Qualitative results obtained for the daytime scene and nighttime scene by fusion method A (U2Fusion~\cite{xu2022u2fusion}) and fusion method B (FBooster~\cite{cheng2025fusionbooster}) on the MSRS dataset.
The ranking change of the mutual information  pixel level feature distributions is presented below these fused images.
}
\label{figure_qualitative_msrs}
\end{figure}

\subsection{Generalisation Experiments on the MSRS dataset}
In this section, we directly apply the proposed EvaNet to the MSRS dataset to evaluate its generalisation capability.
The dataset comprises 361 image pairs captured under both daytime and nighttime conditions, making it well-suited for validating the environment perception capability of EvaNet.

As shown in Fig.~\ref{figure_quantitative_msrs}, EvaNet consistently enhances the $\textrm{MC}_{\textrm{deep}}$ values across multiple image fusion metrics.
With the exception of $\textrm{FMI}_{\textrm{dct}}$, most metrics benefit from improved consistency, indicating that the enhanced versions provided by EvaNet are reliable for evaluating fusion performance on this dataset.
In general, EvaNet achieves at least a 10\% improvement in consistency for several metrics related to human visual perception, while maintaining a significantly faster speed of evaluation.

Fig.~\ref{figure_qualitative_msrs} further illustrates qualitative results produced by two representative fusion methods.
As indicated by the blue circles, during the daytime scene, Method A fails to retain salient texture details from the visible image, resulting in locally under-illuminated regions.
However, the vanilla $\textrm{FMI}_{\textrm{p}}$, which relies solely on pixel-level statistics, does not capture this discrepancy and assigns a higher score to Method A than to Method B, although it handles these regions more effectively.
A similar inconsistency is observed in the nighttime scenario, where the fusion of thermal information conveyed by the infrared modality is evaluated.
Due to the incorporation of the fine-grained quality assessment and environment-aware design, EvaNet mitigates such mismatches by adjusting the metric rankings, thereby reflecting the true perceptual quality of each method's output better.

\begin{table}[tbp]
  \centering
  \caption{The MC value results produced by our EvaNet for reference-free image quality assessments.}
  \resizebox{1\linewidth}{!}{   
    \begin{tabular}{ccccc}
    \hline
    Metric & EN    & SD    & EI    & SF \\
    \hline
    $\textrm{MC}_{\textrm{deep}}$ & 0.659→\textbf{0.679} & 0.715→0.621 & 0.643→0.625 & 0.622→\textbf{0.635} \\
    $\textrm{MC}_{\textrm{clip}}$ & 0.676→0.617 & 0.628→\textbf{0.726} & 0.585→\textbf{0.695} & 0.676→\textbf{0.702} \\
    \hline
    \end{tabular}%
    }
  \label{table_quantitative_noref}%
\end{table}%

\subsection{Improvements to Reference-Free Image Fusion Metrics}
While the proposed EvaNet has been validated on reference-based image fusion metrics, similar limitations, namely, the lack of adaptation to image fusion characteristics, also exist in reference-free metric evaluations.
To further explore this issue, we construct a dedicated dataset to train EvaNet for delivering reference-free quality assessments.
As illustrated in Fig.\ref{figure_qualitative_noreference}, we examine the vanilla version of the entropy (EN) metric by comparing source images and the corresponding fusion results produced by two state-of-the-art methods, Text-IF\cite{yi2024textIF} and GIFNet~\cite{cheng2025gifnet}.
Surprisingly, the infrared source image yields the highest entropy score, suggesting that simply retaining the infrared modality is preferable to fusing it with the visible image.
This counter-intuitive result highlights the inadequacy of using traditional reference-free metrics directly for fusion evaluation.

In contrast, the improved version of the metric (denoted as EN*) addresses this issue by incorporating environment-aware cues and guaging  the texture detail better.
As a result, GIFNet, which delivers more balanced and informative fusion results, receives the top ranking under the adapted metric.

Furthermore, Table~\ref{table_quantitative_ablation} presents the consistency changes of several widely used reference-free metrics after applying EvaNet.
These overall positive trends underscore the potential of EvaNet to serve as a generalisable solution for refining both reference-based and reference-free fusion metrics.

\myrevisedcolor{
\subsection{Limitation and Future Work}
Despite the promising results of EvaNet, several limitations deserve further discussion.
First, the proposed consistency evaluation strategy relies on a fixed set of third-party references, including no-reference quality indicators and downstream task performance, to approximate perceptual alignment. Although objective and reproducible, such proxies cannot fully reflect subjective human preference, and their effectiveness may vary across tasks or model choices.
Second, while this work focuses on infrared and visible image fusion, the underlying consistency issue is not exclusive to IVIF.
For example, in multi-exposure image fusion, many existing evaluation protocols still adopt average-based strategies that implicitly assume equal importance of over- and under-exposed inputs, ignoring scene-dependent contributions.
This limitation is conceptually aligned with the motivation of our environment-aware branch, which aims to dynamically weight heterogeneous inputs based on contextual cues.
Extending EvaNet to multi-exposure fusion therefore represents a natural and promising direction for future research.
Finally, given that our model does not rely on complex source-image computations, it can also be integrated into training pipelines to promote the development of higher-quality image fusion methods.}

\section{Conclusion}
In this paper, we proposed EvaNet, a three-branch learning-based architecture designed for evaluating the quality of the results of infrared and visible image fusion.
Unlike most existing image fusion assessments, which are adapted from general vision tasks without tailored modifications, EvaNet is specifically built to address the unique characteristics of image fusion evaluation.
The proposed model can simultaneously predict multiple widely used fusion metrics, including both reference-based and reference-free variants, with a significantly faster inference speed and a greater consistency with human visual perception.
In contrast to conventional evaluation methods that compute each metric independently, EvaNet delivers $N$ quality assessments in a single forward pass, offering a highly efficient and unified evaluation solution.
Additionally, we introduced a comprehensive evaluation protocol to assess the reliability of any new metric $M$, leveraging high-quality reference models or downstream tasks as supervision.
Extensive experiments conducted across multiple image fusion datasets and downstream benchmarks validate the effectiveness and superiority of our approach.


\section*{Acknowledgement}
This work is supported by the National Key Research and Development Program of China (2023YFE0116300), National Natural Science Foundation of China (62020106012, 62332008, 62106089, 62202205, 62576153), the 111 Project of Ministry of Education of China (B12018), and the Leverhulme Trust Emeritus Fellowship EM-2025-06-09.

\bibliographystyle{IEEEtran}
\bibliography{ref.bib}




\begin{IEEEbiography}
[{\includegraphics[width=1in,height=1.25in,clip,keepaspectratio]{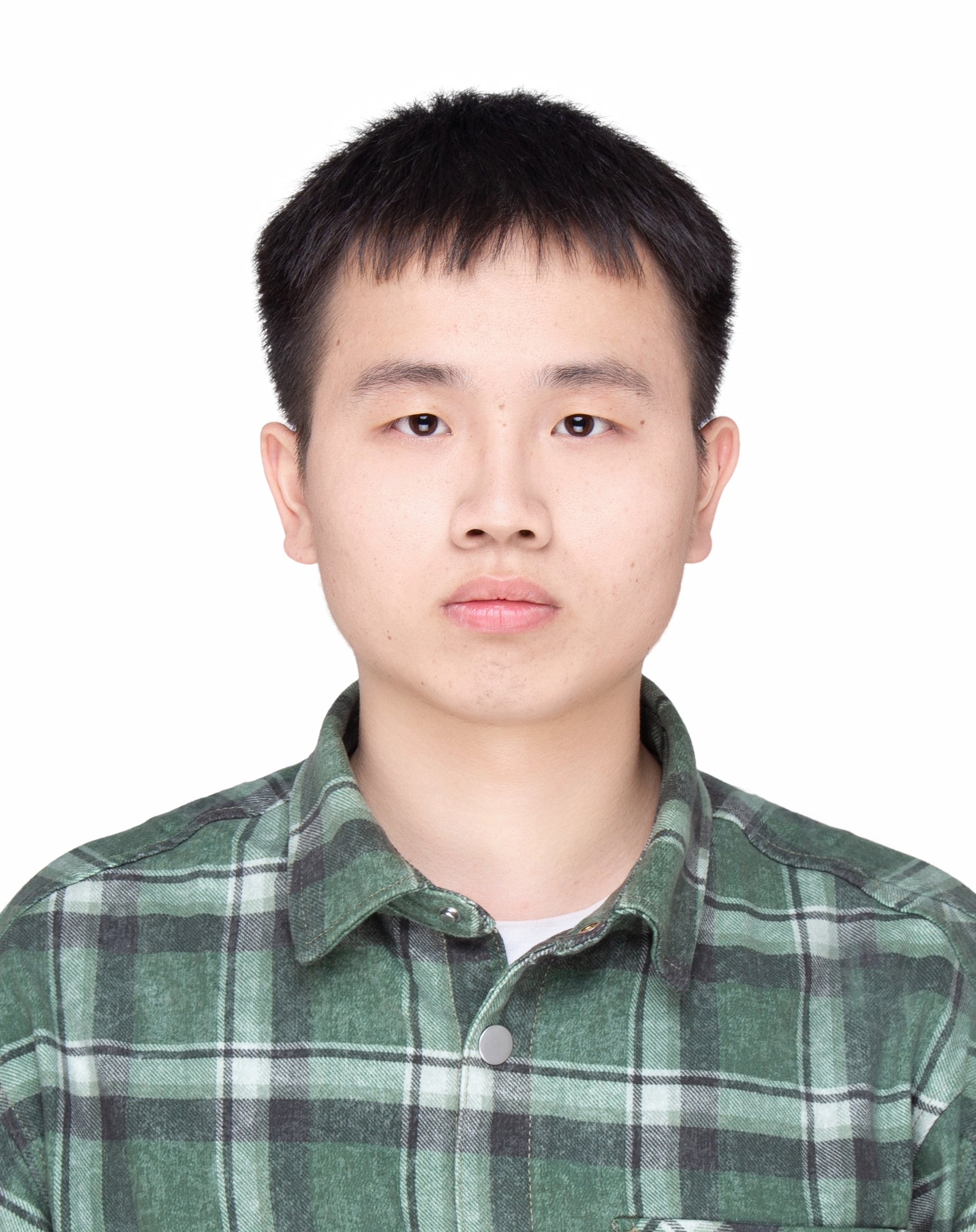}}]
{Chunyang Cheng}
received the Ph.D. degree in Artificial Intelligence and Computer Science from Jiangnan University, Wuxi, China, in 2025. He is currently a Postdoctoral Researcher with the Wuxi School of Medicine and also with the School of Food Science and Technology, Jiangnan University.
He has published several scientific papers, including CVPR, IJCV, IEEE TIM, Information Fusion, IEEE TIP, IEEE TMM, and ACM MM, \textit{etc.}
His research interests include multi-modal fusion and deep learning.
\end{IEEEbiography}
\vspace{-6mm}

\begin{IEEEbiography}
[{\includegraphics[width=1in,height=1.25in,clip,keepaspectratio]{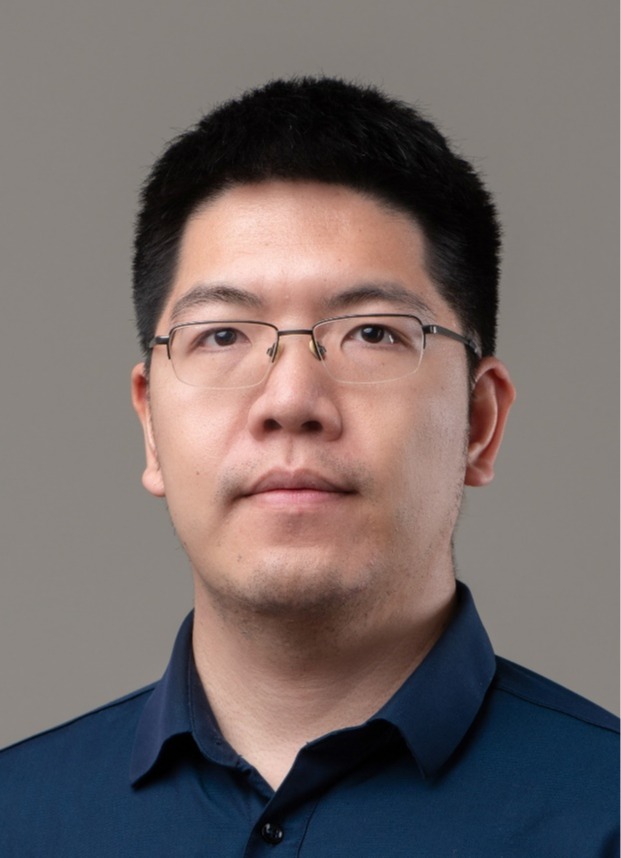}}]
{Tianyang Xu} (Member, IEEE) received the B.Sc. degree in electronic science and engineering from Nanjing University, Nanjing, China, in 2011. He received his Ph.D. degree at the School of Artificial Intelligence and Computer Science, Jiangnan University, Wuxi, China, in 2019. He was a Research Fellow in the Centre for Vision, Speech and Signal Processing (CVSSP), University of Surrey. He is currently an Associate Professor at the School of Artificial Intelligence and Computer Science, Jiangnan University, Wuxi, China. His research interests include computer vision and pattern recognition.
He has published several scientific papers, including IEEE TPAMI, IJCV, IEEE TIP,  CVPR, ICML, ICCV, NeurIPS, ICLR \textit{etc}. He serves as Associate Editor for IEEE TIP and Pattern Recognition. His publications have been cited more than 6000 times. He has been chosen among the World’s Top 2\% Scientists ranking in the single recent year dataset published by Stanford University.
He achieved top 1 performance in several competitions, including the
VOT2020 RGBT challenge (ECCV20), Anti-UAV challenge (CVPR20), and
MMVRAC challenges (ICCV2021, ICME2024).
\end{IEEEbiography}
\vspace{-10mm}

\begin{IEEEbiography}
[{\includegraphics[width=1in,height=1.25in,clip,keepaspectratio]{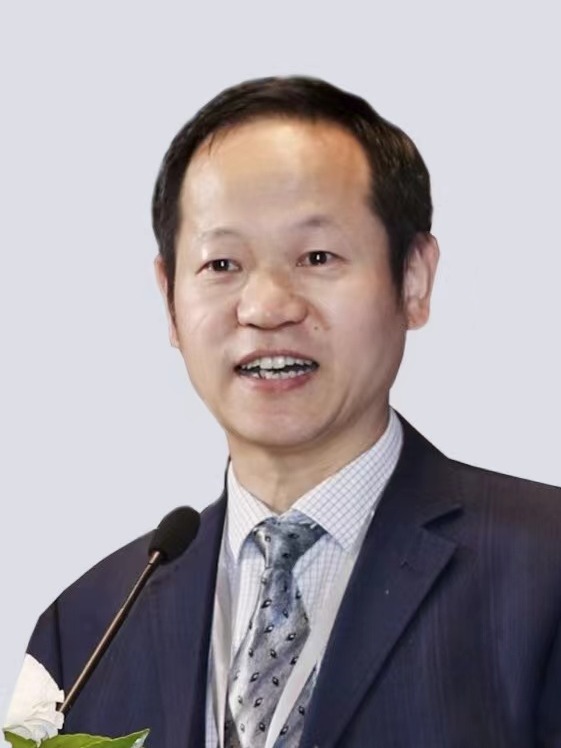}}]
{Xiao-Jun Wu}
(Member, IEEE) received the B.S. degree in mathematics from Nanjing Normal University, Nanjing, China, in 1991, and the M.S. and Ph.D. degrees in pattern recognition and intelligent system from Nanjing University of Science and Technology, Nanjing, in 1996 and 2002, respectively.
He was a fellow of United Nations University, International Institute for Software Technology (UNU/IIST), from 1999 to 2000.
From 1996 to 2006, he taught with the School of Electronics and Information, Jiangsu University of Science and Technology, where he was an Exceptionally Promoted Professor.
In 2006, he joined Jiangnan University, where he is currently a Distinguished Professor with the School of Artificial Intelligence and Computer Science. He was a Visiting Postdoctoral Researcher at the Centre for Vision, Speech, and Signal Processing (CVSSP), University of Surrey, U.K., from 2003 to 2004, under the supervision of Prof. Josef Kittler. He has published more than 400 papers in his fields of research.
His current research interests are pattern recognition, computer vision, fuzzy systems, and neural networks. He received the Most Outstanding Postgraduate Award by Nanjing University of Science and Technology. He owned several domestic and international awards because of his research achievements. He is currently a fellow of IAPR and AAIA.
\end{IEEEbiography}

\begin{IEEEbiography}
[{\includegraphics[width=1in,height=1.25in,clip,keepaspectratio]{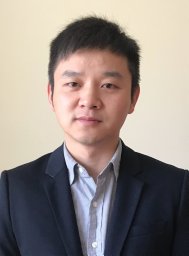}}]
{Tao Zhou} (Senior Member, IEEE) received the Ph.D. degree from Shanghai Jiao Tong University in 2016.
He was a Post-Doctoral Fellow at UNC-CH and a Research Scientist with IIAI.
He is currently an Adjunct Professor at the School of Artificial Intelligence and Computer Science, Jiangnan University, Wuxi, China.
His research interests include machine learning, computer vision, AI in healthcare, and medical image analysis. He is an Associate Editor of IEEE TNNLS, IEEE
TIP, IEEE TMI, and Pattern Recognition.
\end{IEEEbiography}
\vspace{-5mm}

\begin{IEEEbiography}
[{\includegraphics[width=1in,height=1.25in,clip,keepaspectratio]{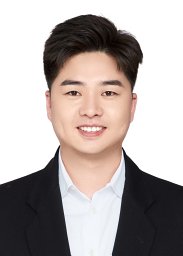}}]
{Hui Li}
(Member, IEEE) received the B.Sc. and
Ph.D. degrees from the School of Internet of Things
Engineering, Jiangnan University, Wuxi, China, in
2015 and 2022, respectively.
He is currently an Associate Professor with the School
of Artificial Intelligence and Computer Science,
Jiangnan University. 
His research interests include image fusion and multi-modal visual information processing. He has been chosen among the World’s Top 2\% Scientists ranking in the single recent year dataset
published by Stanford University (2021–2023). He has published several scientific papers, including IEEE TPAMI, IEEE TIP, and Information Fusion.
He achieved top tracking performance in several competitions, including the VOT2020 RGBT Challenge (ECCV20) and Anti-UAV Challenge (ICCV21).
\end{IEEEbiography}
\vspace{-5mm}

\begin{IEEEbiography}
[{\includegraphics[width=1in,height=1.25in,clip,keepaspectratio]{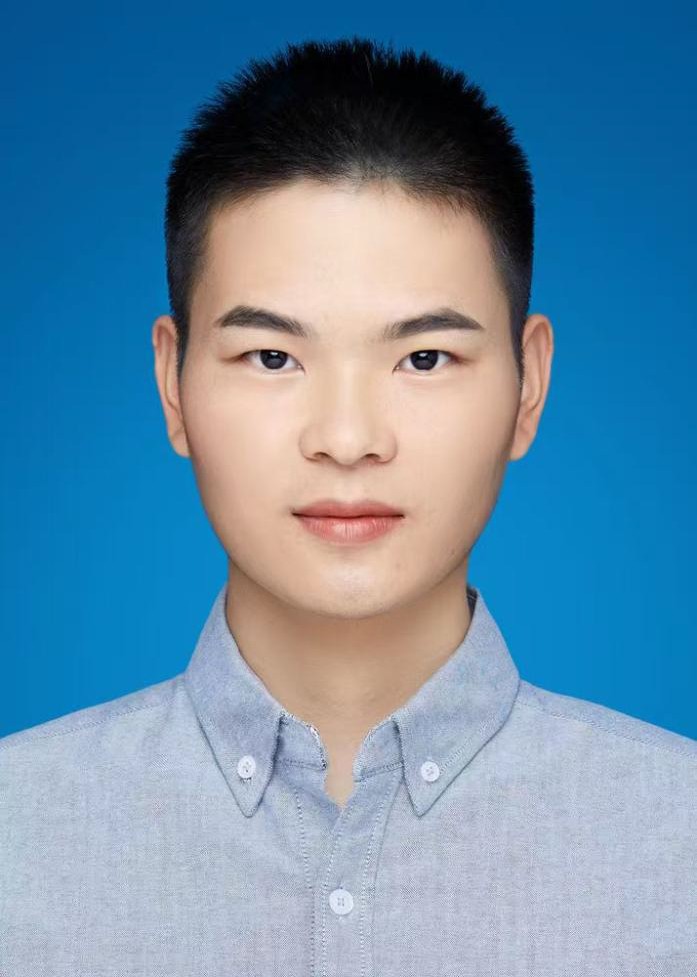}}]
{Zhangyong Tang}
received his Ph.D. degree from Jiangnan University, Wuxi, China, in 2025. He is currently a Postdoc with the The Institute of Translational Medicine, Wuxi School of Medicine, and also with the School of Food Science and Technology, Jiangnan University.
His research interests include multi-modal object tracking and information fusion. He has published several scientific papers, including IJCV, TIP, CVPR, AAAI, ACM MM, \textit{etc}. He achieved top 3 performance in several competitions, including the VOT2020 RGBT challenge (ECCV2020), VOT2021 D challenge (ICCV2021), VOT2022 RGBD challenge (ECCV2022), and the 2\textsuperscript{nd} Anti-UAV challenge.
\end{IEEEbiography}

\begin{IEEEbiography}
[{\includegraphics[width=1in,height=1.25in,clip,keepaspectratio]{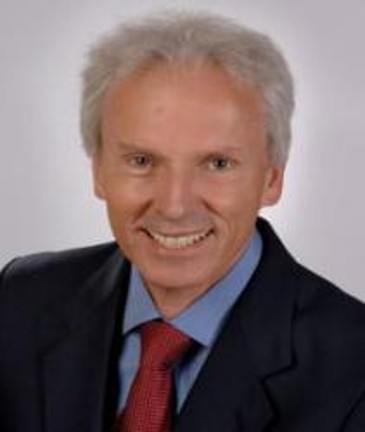}}]
{Josef Kittler} (Life Member, IEEE) received the B.A., Ph.D., and D.Sc. degrees from the University of Cambridge, in 1971, 1974, and 1991, respectively. He is a distinguished professor of machine intelligence with the Centre for Vision, Speech and Signal Processing, University of Surrey, Guildford, U.K. He conducts research in biometrics, video and image database retrieval, medical image analysis, and cognitive vision. He published the textbook Pattern Recognition: A Statistical Approach and about 1000 scientific papers.
His publications have been cited by around 70,000 times. He is series editor of Springer Lecture Notes on Computer Science. He currently serves on the editorial boards of Pattern Recognition Letters, Pattern Recognition and Artificial Intelligence, Pattern Analysis and Applications. He also served as a member of the editorial board of IEEE tpami during 1982–1985.
He served on the Governing Board of the International Association for Pattern Recognition (IAPR) as one of the two British representatives during 1982–2005, and the president of IAPR during 1994–1996.
\end{IEEEbiography}

\vfill

\end{document}